\documentclass[11pt]{article}

\usepackage[T1]{fontenc}
\usepackage[margin=1in]{geometry}
\usepackage{amsmath,amssymb,amsthm,mathtools}
\usepackage{lmodern}
\usepackage{microtype}
\usepackage{enumitem}
\usepackage{float}
\usepackage{needspace}
\usepackage{xcolor}
\usepackage[numbers,sort&compress]{natbib}
\usepackage[colorlinks=true,linkcolor=blue!55!black,citecolor=blue!55!black,urlcolor=blue!55!black]{hyperref}
\usepackage[nameinlink,capitalise,noabbrev]{cleveref}

\usepackage{latexsym,tikz,graphicx}
\usetikzlibrary{calc}
\usetikzlibrary{decorations.pathreplacing}
\usetikzlibrary{patterns}

\hypersetup{
  pdftitle={Tight Mistake–Abstention Tradeoffs for Sequential Prediction under Adversarial Injections},
  pdfauthor={Anonymous},
  pdfsubject={Online learning with abstention under adversarial injections}
}

\allowdisplaybreaks
\setlist[itemize]{leftmargin=2em,itemsep=0.2em,topsep=0.3em}
\setlist[enumerate]{leftmargin=2.2em,itemsep=0.2em,topsep=0.3em}

\newtheorem{theorem}{Theorem}[section]
\newtheorem{lemma}{Lemma}[section]
\newtheorem{proposition}{Proposition}[section]
\newtheorem{corollary}{Corollary}[section]
\theoremstyle{definition}
\newtheorem{definition}{Definition}[section]
\theoremstyle{remark}
\newtheorem{remark}{Remark}[section]

\floatstyle{ruled}
\newfloat{algorithm}{tbp}{loa}
\floatname{algorithm}{Algorithm}
\newenvironment{algorithmic}[1][]{%
  \begin{enumerate}[leftmargin=3.0em,label=\arabic*:,itemsep=0.08em,topsep=0.2em]
}{%
  \end{enumerate}
}
\newcommand{\State}{\item}

\newcommand{\Require}{\item \textbf{Require:}~}
\newcommand{\For}[1]{\item \textbf{for} #1 \textbf{do}\begin{enumerate}[leftmargin=1.8em,label={},itemsep=0.08em,topsep=0.1em]}
\newcommand{\EndFor}{\end{enumerate}}
\newcommand{\If}[1]{\item \textbf{if} #1 \textbf{then}\begin{enumerate}[leftmargin=1.8em,label={},itemsep=0.08em,topsep=0.1em]}
\newcommand{\Else}{\end{enumerate}\item \textbf{else}\begin{enumerate}[leftmargin=1.8em,label={},itemsep=0.08em,topsep=0.1em]}
\newcommand{\EndIf}{\end{enumerate}}
\newcommand{\Comment}[1]{\hfill$\triangleright$~\textit{#1}}

\newcommand{\X}{\mathcal X}
\newcommand{\F}{\mathcal F}
\newcommand{\V}{\mathcal V}

\newcommand{\E}{\mathbb E}
\newcommand{\Pp}{\mathbb P}
\newcommand{\R}{\mathbb R}
\newcommand{\1}{\mathbf 1}

\newcommand{\MisErr}{\mathsf{MisErr}}
\newcommand{\AbsErr}{\mathsf{AbsErr}}

\newcommand{\VC}{\operatorname{VC}}
\newcommand{\Sh}{\operatorname{Sh}}
\newcommand{\Unif}{\operatorname{Unif}}

\newcommand{\poly}{\operatorname{poly}}

\newcommand{\shatters}{\mathrel{\triangleright}}
\DeclareMathOperator*{\argmax}{arg\,max}

\newcommand{\paren}[1]{\left( #1 \right)}
\newcommand{\sqb}[1]{\left[ #1 \right]}
\newcommand{\set}[1]{\left\{ #1 \right\}}

\newcommand{\ceil}[1]{\left\lceil #1 \right\rceil}
\newcommand{\abs}[1]{\left|#1\right|}

\newcommand{\Ocal}{\mathcal O}
\newcommand{\Ncal}{\mathcal N}

\newcommand{\comment}[1]{}
\definecolor{darkgreen}{RGB}{0,100,0}      
\definecolor{forestgreen}{RGB}{40,180,40}  

\usepackage{subcaption}
 
\tikzset{
  tradeoff plot/.style={x=3.2cm, y=3.2cm, line cap=round},
  ruled out/.style={fill=red!40},
  achieved/.style={fill=forestgreen!40},
  lower bound/.style={red, ultra thick},
  upper bound/.style={forestgreen, ultra thick},
  regime boundary/.style={dashed, thin, black!55},
  included/.style={fill=forestgreen},
  excluded/.style={draw=forestgreen, fill=white, thick},
  curve label/.style={draw=forestgreen, fill=white, text=forestgreen,
                      rounded corners=1pt, inner sep=1.5pt, font=\small\bfseries},
}
 
\crefname{algorithm}{Algorithm}{Algorithms}
\crefname{theorem}{Theorem}{Theorems}
\crefname{lemma}{Lemma}{Lemmas}
\crefname{proposition}{Proposition}{Propositions}
\crefname{corollary}{Corollary}{Corollaries}
\crefname{definition}{Definition}{Definitions}

\title{Tight Mistake–Abstention Tradeoffs for Sequential Prediction under Adversarial Injections}

\author{
Mo\"ise Blanchard$^{1}$\thanks{Authors are listed in alphabetical order.
Emails: \texttt{\{mblanchard41, yujl1\}@gatech.edu},
\texttt{\{ezrae, surbhig\}@cis.upenn.edu},
\texttt{shyamsn@mit.edu}.}
\and
Ezra Edelman$^{2}$
\and
Surbhi Goel$^{2}$
\and
Shyam Narayanan$^{3}$
\and
Jialin Yu$^{1}$
\\[1ex]
\small
$^{1}$Georgia Institute of Technology
\qquad
$^{2}$University of Pennsylvania
\qquad
$^{3}$Citadel
}

\date{}

\begin{document}
\newcommand{\trw}{\text{\small TRW}}
\newcommand{\maxcut}{\text{\small MAXCUT}}
\newcommand{\maxcsp}{\text{\small MAXCSP}}
\newcommand{\suol}{\text{SUOL}}
\newcommand{\wuol}{\text{WUOL}}
\newcommand{\crf}{\text{CRF}}
\newcommand{\sual}{\text{SUAL}}
\newcommand{\suil}{\text{SUIL}}
\newcommand{\fs}{\text{FS}}
\newcommand{\fmv}{{\text{FMV}}}
\newcommand{\smv}{{\text{SMV}}}
\newcommand{\wsmv}{{\text{WSMV}}}
\newcommand{\trwp}{\text{\small TRW}^\prime}
\newcommand{\alg}{\text{ALG}}
\newcommand{\rhos}{\rho^\star}
\newcommand{\brhos}{\brho^\star}
\newcommand{\bzero}{{\mathbf 0}}
\newcommand{\bs}{{\mathbf s}}
\newcommand{\bw}{{\mathbf w}}
\newcommand{\bws}{\bw^\star}
\newcommand{\ws}{w^\star}
\newcommand{\Prt}{{\mathsf {Part}}}
\newcommand{\Fs}{F^\star}

\newcommand{\Hs}{{\mathsf H} }

\newcommand{\hL}{\hat{L}}
\newcommand{\hU}{\hat{U}}
\newcommand{\hu}{\hat{u}}

\newcommand{\bu}{{\mathbf u}}
\newcommand{\ubf}{{\mathbf u}}
\newcommand{\hbu}{\hat{\bu}}

\newcommand{\primal}{\textbf{Primal}}
\newcommand{\dual}{\textbf{Dual}}

\newcommand{\Ptree}{{\sf P}^{\text{tree}}}
\newcommand{\bv}{{\mathbf v}}

\newcommand{\bq}{\boldsymbol q}

\newcommand{\rvM}{\text{M}}

\newcommand{\Acal}{\mathcal{A}}
\newcommand{\Bcal}{\mathcal{B}}
\newcommand{\Ccal}{\mathcal{C}}
\newcommand{\Dcal}{\mathcal{D}}
\newcommand{\Ecal}{\mathcal{E}}
\newcommand{\Fcal}{\mathcal{F}}
\newcommand{\Gcal}{\mathcal{G}}
\newcommand{\Hcal}{\mathcal{H}}
\newcommand{\Ical}{\mathcal{I}}
\newcommand{\Kcal}{\mathcal{K}}
\newcommand{\Lcal}{\mathcal{L}}
\newcommand{\Mcal}{\mathcal{M}}
\newcommand{\Pcal}{\mathcal{P}}
\newcommand{\Scal}{\mathcal{S}}
\newcommand{\Tcal}{\mathcal{T}}
\newcommand{\Ucal}{\mathcal{U}}
\newcommand{\Vcal}{\mathcal{V}}
\newcommand{\Wcal}{\mathcal{W}}
\newcommand{\Xcal}{\mathcal{X}}
\newcommand{\Ycal}{\mathcal{Y}}
\newcommand{\Qcal}{\mathcal{Q}}
\newcommand{\Rcal}{\mathcal{R}}

\newcommand{\brho}{\boldsymbol{\rho}}

\newcommand{\Cbb}{\mathbb{C}}
\newcommand{\Ebb}{\mathbb{E}}
\newcommand{\Nbb}{\mathbb{N}}
\newcommand{\Pbb}{\mathbb{P}}
\newcommand{\Qbb}{\mathbb{Q}}
\newcommand{\Rbb}{\mathbb{R}}
\newcommand{\Sbb}{\mathbb{S}}
\newcommand{\Vbb}{\mathbb{V}}
\newcommand{\Wbb}{\mathbb{W}}
\newcommand{\Xbb}{\mathbb{X}}
\newcommand{\Ybb}{\mathbb{Y}}
\newcommand{\Zbb}{\mathbb{Z}}

\newcommand{\Rbbp}{\Rbb_+}

\newcommand{\bX}{{\mathbf X}}
\newcommand{\bx}{{\boldsymbol x}}

\newcommand{\btheta}{\boldsymbol{\theta}}

\newcommand{\Pb}{\mathbb{P}}

\newcommand{\hPhi}{\widehat{\Phi}}

\newcommand{\Sigmah}{\widehat{\Sigma}}
\newcommand{\thetah}{\widehat{\theta}}

\newcommand{\notindep}{\not\!\perp\!\!\!\perp}

\newcommand{\one}{\mathbbm{1}}
\newcommand{\aprx}{\alpha}

\newcommand{\ST}{\Tcal(\Gcal)}
\newcommand{\x}{\mathsf{x}}
\newcommand{\y}{\mathsf{y}}
\newcommand{\Ybf}{\textbf{Y}}
\newcommand{\smiddle}[1]{\;\middle#1\;}
\newcommand{\ind}{\1}

\definecolor{dark_red}{rgb}{0.2,0,0}
\newcommand{\detail}[1]{\textcolor{dark_red}{#1}}

\newcommand{\ds}[1]{{\color{red} #1}}
\newcommand{\rc}[1]{{\color{green} #1}}

\newcommand{\mb}[1]{\ensuremath{\boldsymbol{#1}}}

\newcommand{\todo}[1]{{\color{red} TODO: #1}}
\newcommand{\algoname}{\textsc{AbstainBoost} }
\maketitle

\begin{abstract}
We study the sequential prediction problem introduced by
\citet{goel2023adversarial}, where an adversary can inject arbitrarily many corrupted instances in an i.i.d. stream from a clean distribution $\mu$, while the learner is allowed to \emph{abstain} from making a prediction at no cost if the current instance was indeed injected. When $\mu$ is unknown, \citet{yu2026distributionfreev2} show a polynomial tradeoff between mistakes and false abstentions for classes with finite VC dimension and oblivious adversaries, but still leave open a significant gap between upper and lower bounds in their tradeoff landscape. In this paper, we propose the weighted shattering potential, and based on variants of this potential, provide gap-closing algorithms that essentially characterize the tradeoffs under various settings of interest: When $\mu$ is known, our algorithm improves over \cite{goel2023adversarial} and achieves their conjectured $O(d\log T)$ mistake bound; When $\mu$ is unknown, for oblivious adversary, our guarantees are tight both in $T$ and in terms of the VC dimension $d$, which completely characterizes the abstention-mistake tradeoff in this setting and gives an $O(\sqrt{dT})$ balanced rate. For adaptive adversary, our results are tight up to $\poly(dD\log T)$ factors for structured classes with finite so-called reduction dimension $D$. Our guarantees also extend to \emph{censored} feedback, where the label is hidden after an abstention.
\end{abstract}

\section{Introduction}
\label{sec:introduction}

Consider the sequential prediction problem where a learner sequentially observes each instance and predicts on its label before receiving its true label. 
We will focus on the realizable setting, where the labels are generated by an unknown target $f^\star$ in a hypothesis class $\F$. 
The following two assumptions on the sequence of instances are standard in the literature. In the stochastic setting, the instances are drawn i.i.d.\ from a fixed unknown distribution, whereas in the fully adversarial setting, the instances may be chosen adaptively by an adversary depending on the learner and the past instances.  These two settings can have very different complexities. In particular, finite VC
dimension is enough for learning in the stochastic setting \citep{vapnik1971uniform,vapnik1974theory,valiant1984theory}, whereas a finite worst-case mistake bound on an adversarial sequence requires finite Littlestone dimension
\citep{littlestone1988learning}, which does not even hold for thresholds on the real line. 

The clean-label injection model of \citet{goel2023adversarial} lies between these two extremes. In this model, there is an underlying stream of i.i.d. examples from a fixed, unknown clean distribution $\mu$, but an
adversary is allowed to inject an arbitrary number of corrupted instances into the stream. The learner observes this combined stream and does not know whether the instances are clean or injected. 

The learner is allowed to abstain on instances, and its error is measured by the number of mistakes on rounds that the learner makes a prediction, and the number of
abstention it makes on clean instances.  Thus the learner can safely abstain on
an injected instances without any penalty. 

When the clean distribution is known, \citet{goel2023adversarial} gave a learner that makes
$O(d^2\log T)$ mistakes and $O(d)$ expected clean abstentions for every class of VC dimension
$d$. On the other hand, when the distribution is unknown and the adversary is oblivious, \citet{yu2026distributionfreev2} gave a learner that makes $O(T^{3/4})$ mistakes and
$O(d^2(\log T)^{5/3}T^{3/4})$ expected clean abstentions. While these results show that the complexity of learning in this model is governed by the VC dimension, it is unknown if these are the optimal trade-offs between mistakes and abstentions. This leads to the main question of the paper: 

\begin{center}
    \emph{What are the optimal mistake--abstention tradeoffs for sequential prediction\\under
adversarial injections?}
\end{center}

\subsection{Main Contributions}
\label{sec:contributions}

We summarize our main results below.  We write $\MisErr$ for the total number of mistakes and
$\AbsErr$ for the number of abstentions on clean rounds. All expectation bounds average over the clean draws and any randomization of the learner or
adversary.  

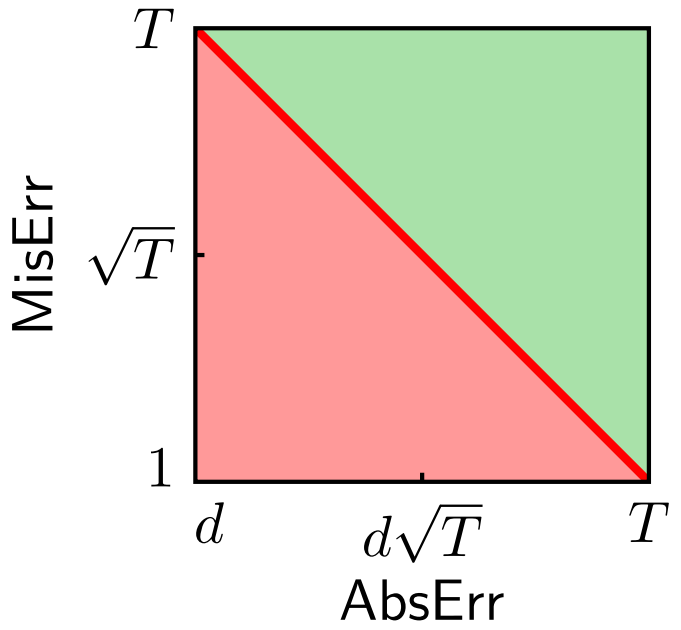
\begin{figure}[t]
  \centering
    \begin{tikzpicture}[scale=3]
      \begin{scope}
        \clip (0,0) rectangle (1,1);
        \fill[ruled out] (0,0) -- (1,0) -- (0,1) -- cycle;
        \fill[achieved]  (0,1) -- (1,1) -- (1,0) -- cycle;
        \draw[lower bound] (0,1) -- (1,0);
      \end{scope}
        \draw[thick] (0,0) rectangle (1,1);
      \foreach \pos in {0.5}{
        \draw[thick] (\pos,0) -- (\pos,0.02);
        \draw[thick] (0,\pos) -- (0.02,\pos);
      }
      \node[below] at (0.03,0) {$d$};
      \node[below] at (0.5,0)  {$d\sqrt{T}$};
      \node[below] at (1,0)    {$T$};
      \node[below] at (0.5,-0.17) {$\AbsErr$};
      \node[left] at (0,0.03) {$1$};
      \node[left] at (0,0.5)  {$\sqrt{T}$};
      \node[left] at (0,1)    {$T$};
      \node[rotate=90, anchor=south] at (-0.27,0.5) {$\MisErr$};
    \end{tikzpicture}
    \caption{This is a schematic summary of the tradeoffs between $\AbsErr$ and $\MisErr$ that we achieve in this paper. The plot is exact for VC-dimension-$d$
    classes under oblivious injections (with full or censored feedback) when
    $d<\sqrt{T}$.  Axes are log-scale in $T$.  The red region is ruled out by \cref{thm:unknown-lower-bound}. The green region is achieved by \cref{thm:unknown-oblivious-no-log,thm:censored-age-restarts} under full and censored feedback respectively, giving a complete characterization. For adaptive injections, the lower bound still applies, and we achieve a similar upper bound in \cref{thm:adaptive-full-feedback-direct}, tight up to $\text{poly}(dD\log T)$ factors, where $D$ is the $O(d\log T)$-reduction dimension of the class.}
  \label{fig:tradeoffs}
\end{figure}

\begin{itemize}[leftmargin=1.8em]
  \item When the clean distribution is known, we give a learner that works against adaptive
  injections and satisfies, for every $\beta>0$,
  \[
    \MisErr\leq \frac{d\log(1+\beta)}{\log(4/3)}
    \qquad\text{and}\qquad
    \E[\AbsErr]\leq \frac{2T}{\beta}.
  \]
  The mistake bound holds deterministically on every realized interaction.  Taking $\beta=T$ gives
  $O(d\log T)$ mistakes and at most two expected clean abstentions, resolving the question of
  \citet{goel2023adversarial}.

  \item When the clean distribution is unknown and the injections are oblivious, for every $\alpha\in[0,1)$ we give a
  learner for every finite-VC class with
  \[
    \MisErr=O(T^\alpha)
    \qquad\text{and}\qquad
    \E[\AbsErr]=O(dT^{1-\alpha}).
  \]
  Again, the mistake bound holds on every realized interaction. For censored feedback---where the label is hidden after an abstention---we also construct a learner essentially achieving the same guarantee up to constant factors.
  We complement this with a matching lower bound: for each fixed $\alpha\in[0,1)$ and $d\geq1$, there exists a class of VC dimension exactly $d$ and an oblivious clean-label process such that every learner incurs either
    \[
    \E[\MisErr]=\Omega(T^\alpha)
    \qquad\text{or}\qquad
    \E[\AbsErr]
    =\Omega\!\left(\min\{dT^{1-\alpha},T\}\right),
    \]
    even under full feedback. Thus the lower bound matches the upper tradeoff throughout the full parameter range. Our findings are summarized in \cref{fig:tradeoffs}.

  \item Last, when the injections are adaptive, we improve the mistake/abstention-error tradeoffs from \cite{yu2026distributionfreev2} for classes with finite reduction dimension. This result uses a similar candidate-bank reduction, with (1) the weighted potential supplying a sharper comparator (2) a sample-efficient score compression that largely reduces the number of experts and (3) a sharper expert aggregation strategy based on multiplicative weights.
  Combining the three leads to the following guarantee: Assuming the function class has $O(d\log T)$-reduction dimension $D$, then for $\alpha \in [0,1]$, we give a learner such that
  \[
    \MisErr = O (T^\alpha + d^2 D \log^4 T)\qquad\text{and}\qquad \E[\AbsErr] = O(d^2 D\,T^{1-\alpha}\log^4 T),
\]
  even under censored feedback.
  In particular, for linear classifiers in $\Rbb^p$, we have
  \[
    \MisErr=O(T^{\alpha}+p^5 \log^5 T)\qquad\text{and}\qquad\E[\AbsErr]=O(p^5 T^{1-\alpha} \log^5 T).
  \]
\end{itemize}

As a remark, our algorithms for the unknown clean distribution and oblivious adversary (uncensored case) only rely on exchangeability of clean samples. Hence, their guarantees still hold under a stronger model where the adversary may also select all clean samples but these arrive in a random order within clean rounds. We refer to \cref{remark:exchangeable_setting} for a formal description of this more general learning model.

\subsection{Technical Overview: A Weighted Shattering Potential}
\label{sec:proof-overview}

\paragraph{A weighted shattering potential when the clean distribution is known.} A central idea is a single measure of the uncertainty remaining in the current version space.
Suppose first that the clean distribution is known.  For a version space $V$, let
$\rho_k(V)$ be the probability that $V$ shatters $k$ independent clean draws.  Earlier
work follows one value of $k$ at a time.  At level $k$, a mistake reduces
$\rho_k(V)$ by a constant factor; when this quantity becomes too small, the analysis moves to
level $k-1$ and starts a new progress budget.  The budget at level $k$ is
$O(k\log T)$, and summing these budgets is what produces the quadratic dependence on $d$.

We instead combine all levels in the weighted potential
\[
  \Theta_\beta(V):=\sum_{k=0}^d \beta^k\rho_k(V).
\]
The geometric weights are chosen to match the relation between consecutive shattering levels.
When the current point is a fresh clean draw, the contribution obtained by extending a shattered
$k$-tuple has expectation $\rho_{k+1}(V)$.  Its weight at level $k$ is
$\beta^k\rho_{k+1}(V)$, exactly $1/\beta$ times the corresponding term at level $k+1$.
Thus all levels can be charged to the same potential, rather than to $d$ separate budgets.

On each round, the learner computes the potential after the two possible label restrictions.  It
abstains if both restricted classes remain large and otherwise predicts according to the larger
one.  A mistake sends the version space to the smaller branch and contracts the potential by a
constant factor.  Since the potential starts below $(1+\beta)^d$ and never falls below one,
there can be only $O(d\log(1+\beta))$ mistakes.  The same one-point calculation shows that both
branches are large on a clean draw with probability at most $2/\beta$, giving the abstention
bound.

\paragraph{An empirical potential for the unknown distribution case.} When the clean distribution is unknown, the probabilities $\rho_k(V)$ cannot be evaluated from
the contaminated history.  We keep the same geometric weighting but apply it directly to
shattered subsets of the observed labeled sequence. This deterministic potential is not an
estimate of the clean distribution.  Its permutation symmetry, together with exchangeability of
the clean examples, is what yields the unknown-distribution result. Using a direct empirical analog of the potential $\Theta_\beta(V)$ would already give the correct misclassification/abstention tradeoff up to a logarithmic factor. To remove these logarithmic factors, we modify the weight $\beta^k$ for $k$-shattered sets $U$ to take into account the ``age'' of datapoints in $U$. Intuitively, we incentivize the learner to resolve older datapoints and focus on more recent information.

\paragraph{Censored potential and control of version space.} Adjusting the potential to the censored case requires a few modifications: potentially the learner does not observe labels on most rounds. Fortunately, the contexts $x_t$ are always available and can be used to compute shattered sets of version spaces induced by labeled data. However, the version spaces encountered throughout the learning procedure depend on the history, potentially breaking the previous exchangeability argument between clean samples. To resolve this issue, we use a uniform concentration argument over all possible version spaces. To control the number of version spaces, we only update the version space when the algorithm makes misclassification errors, and we periodically restore the version space to its initialization.

\paragraph{Compressed witnesses and direct aggregation for adaptive injections.} We build on the candidate-bank approach of \cite{yu2026distributionfreev2}: although the learner cannot identify clean observations, it aggregates a family of experts containing one whose bank consists entirely of clean samples. Our reference expert predicts using averages of normalized weighted-shattering potentials over bank blocks. Each mistake contracts the potential, while a leave-one-out argument, made uniform over constrained version spaces using reduction dimension, controls clean abstentions. To reduce the size of the candidate family, we compress each prediction score into a short list of shattered witness tuples whose description length is much smaller than the bank size. 
Finally, we use a direct multiplicative-weight-based approach to aggregate these compressed experts, which penalizes experts making mistakes while rewarding those making predictions.

\subsection{Related Work}
\label{sec:related-work}

\paragraph{Adversarial injections.}
The closest work is that of \citet{goel2023adversarial}, who introduced the model.  For a known
clean distribution, their algorithm follows one shattering level at a time and obtains a quadratic
dependence on the VC dimension.  For an unknown distribution, they give algorithms for
VC-dimension-one classes and axis-aligned rectangles, and suggest leave-out shattering statistics
as a possible route to general classes.  Our known-distribution learner combines all shattering
levels in one potential.  Our unknown-distribution learner uses the same kind of leave-out
witnesses, but counts them deterministically instead of treating them as estimates of a population
probability.

\citet{yu2026distributionfreev2} obtained the first unknown-distribution guarantee for every
finite-VC class under an oblivious adversary.  Their construction builds many candidate labeled banks and aggregates the
resulting learners so that one candidate is based on clean data.  They also develop reductions for
adaptive injections and censored feedback.  We use these reductions in our corresponding
extensions, with the weighted learner as a sharper base algorithm.  Our oblivious full-feedback
and censored learners do not require candidate-bank aggregation.

\citet{edelman2026reliable} take a different approach
based on robust witnesses: a small labeled subset certifies a prediction and forces progress even
when the history is contaminated. They obtain class-specific upper bounds and an
$\Omega(\sqrt T)$ lower bound at VC dimension one.  We extend the lower bound to VC dimension $d$ classes. Also, our algorithm follows essentially the template of their potential-based algorithm with the potential instantiated as a weighted shattering. 

\paragraph{Transductive abstention under distribution shift.}
The offline analogue of the online adversarial model was introduced by
\citet{goldwasser2020identifying}. In their PQ learning model, the learner is given an i.i.d.\ labeled training sample and an unlabeled test set, some of which may be inserted/modified by an adversary.  The goal is to output a predictor and selector such that the error on the test and false abstentions on the training are simultaneously minimized.  \citet{kalai2021optimally} also assume transductive access and study
the optimal tradeoff when every abstention has a fixed cost. \citet{klivans2024testable} introduced the decision version of this problem called Testable Distribution Shift (TDS) learning where the learner may abstain on the entire test set with the motivation that the decision problem would allow for computationally efficient algorithms, with subsequent advances for intersections of halfspaces, discrepancy testing, and neural-network regression \citep{klivans2024intersections,chandrasekaran2024discrepancy,chandrasekaran2025neural}. Surprisingly, recent work by \citet{patel2026equivalence} showed that these two models of learning are equivalent in the distribution free setting.

\paragraph{Other abstention objectives.}
Classical selective classification studies the error--coverage tradeoff under a single test distribution \citep{chow1970optimum,elyaniv2010foundations}.  The online learning versions either charge every abstention or combine mistakes and abstentions into one objective \citep{zhang2016extended,cortes2018online,neu2020fast}.  The model is different in two ways: prediction errors are counted on both clean and injected rounds, and only clean abstentions are charged.  This hidden distinction between the two sources is what allows the statistical clean subsequence to help without revealing which inputs came from it.
\section{Setup}
\label{sec:model}


We consider the sequential online learning model with clean-label corruptions introduced by \citet{goel2023adversarial}. Throughout, we denote by $\Xcal$ the instance space and let $\Fcal\subseteq\{0,1\}^{\Xcal}$ be a hypothesis class. In the classical sequential prediction problem---without any corruptions---the VC dimension exactly characterizes the learnability of a function class \citep{vapnik1971uniform,vapnik1974theory,valiant1984theory}.
\begin{definition}[Shattering and VC dimension] Let $\Fcal\subseteq\{0,1\}^\Xcal$ be a function class on $\Xcal$. We say that $\Fcal$ \emph{shatters} a set of points $x_1,...,x_k\in\Xcal$ if
for any choice of labels $\epsilon\in\{0,1\}^k$ there exists $f\in\Fcal$ such that $f(x_i)=\epsilon_i$ for all $i\in[k]$. The VC dimension of $\Fcal$ is the size of the
largest shattered set.
\end{definition}
Accordingly, we aim to achieve learning guarantees in the abstention learning setting when the function class $\Fcal$ has finite VC dimension, denoted by $d$. For adaptive adversaries, we will also consider classes with finite so-called reduction-dimension as defined in \cite{yu2026distributionfree}.

\begin{definition}[$\ell$-reduction dimension \citep{yu2026distributionfree}]
\label{def:adaptive-reduction-dimension}
For $\ell\ge1$, the class $\mathcal F$ has $\ell$-reduction dimension $D_\ell\ge1$ if,
for every $n\ge D_\ell$ and every $x_1,\ldots,x_n\in\mathcal X$,
\[
  \left|
    \left\{
      (\mathcal F\cap A)|_{\{x_1,\ldots,x_n\}}:
      A\subseteq\mathcal X\times\{0,1\},\ |A|\le\ell
    \right\}
  \right|
  \le
  n^{D_\ell},
\]
where $\mathcal F\cap A:=\{f\in\mathcal F:f(x)=y\text{ for every }(x,y)\in A\}$.
\end{definition}

The learning protocol is as follows.

\paragraph{Learning protocol.} Before the interaction begins, an adversary selects a clean instance distribution $\mu$ on $\Xcal$ together with an unknown target $f^\star\in\Fcal$ to the learner. Then, at each round of interaction, the adversary chooses whether to inject a corrupted instance or draw a clean sample. The learner then sees the resulting instance and may choose to predict a label or abstain. Formally, at each round $t\in[T]$:
\begin{enumerate}
  \item The adversary decides whether to corrupt the round or not by selecting $c_t\in\{0,1\}$.
  \item If $c_t=0$, the round is \emph{clean} and the new instance is a fresh draw from the clean distribution $x_t\sim\mu$. If $c_t=1$, the round is \emph{corrupted} and the adversary chooses $x_t\in\Xcal$ arbitrarily.
  \item After observing $x_t$, the learner selects
        $\widehat y_t\in\{0,1,\bot\}$, where $\bot$ denotes an abstention.
  \item The true label is $y_t=f^\star(x_t)$, and the feedback rule determines whether the learner observes it.
\end{enumerate}
Importantly, the learner never observes $c_t$, i.e., whether the round is clean or corrupted. Note that while we allow for arbitrary corruptions in instances, labels remain realizable and consistent with the fixed target $f^\star$.

\comment{
Let $\X$ be the instance space and let
$\F\subseteq\{0,1\}^{\X}$ be a hypothesis class with finite VC dimension
$d:=\VC(\F)\geq 1$.  
Before the interaction begins, fix a horizon
$T\in\mathbb N$, a clean distribution $\mu$ on $\X$, and a target
$f^\star\in\F$.  All guarantees are uniform over the pair $(\mu,f^\star)$, and the adversary
may know both.  The learner never knows $f^\star$; it is told $\mu$ only in results explicitly
designated as known-distribution.  

The round begins with a hidden choice between a clean draw and an injection.  The learner then sees
the resulting instance, chooses whether to predict, and receives feedback according to the
specified feedback model.  Formally, each round $t\in[T]$ proceeds as follows:
\begin{enumerate}
  \item the adversary chooses a corruption indicator $c_t\in\{0,1\}$;
  \item if $c_t=0$, the instance $x_t$ is a fresh draw from $\mu$; if $c_t=1$, the
        adversary chooses $x_t\in\X$;
  \item after observing $x_t$, the learner outputs
        $\widehat y_t\in\{0,1,\bot\}$;
  \item the label is $y_t=f^\star(x_t)$, and the feedback rule determines whether the learner
        observes it.
\end{enumerate}
Rounds with $c_t=0$ are \emph{clean}; rounds with $c_t=1$ are \emph{corrupted}, or
\emph{injected}.  
}

The goal of the learner is to make few prediction mistakes while not abstaining on clean rounds, that is, mistakes are counted on every round, whereas abstentions are charged only on clean rounds. Formally, the misclassification and abstention errors are respectively defined as 
\begin{align*}
  \MisErr
  &:=\sum_{t=1}^T\1\{\widehat y_t\in\{0,1\},\ \widehat y_t\neq y_t\},\\
  \AbsErr
  &:=\sum_{t=1}^T\1\{c_t=0,\ \widehat y_t=\bot\}.
\end{align*}
As a remark, while the learner can observe its misclassification error, it never observes $c_t$ and hence neither observes which abstentions were charged.

\paragraph{Full vs.\ censored feedback.} We consider two different feedback mechanisms.
Under \emph{full feedback}, $y_t$ is revealed after every round including those on which the
learner abstains. In contrast, under \emph{censored feedback}, $y_t$ is revealed only when a prediction is made:
$\widehat y_t\in\{0,1\}$.

\paragraph{Rationale of the model.}
A motivating example is speculative execution. A learner predicts a
program's binary branch outcome and begins executing the predicted branch,
or abstains and waits for the condition to resolve. Evaluating the condition
naturally reveals the correct label. A correct prediction allows useful
work to begin earlier, whereas an incorrect prediction incurs wasted compute.

The objective of learner is to accelerate ordinary workloads while avoiding slowdowns
on injected inputs relative to conventional, non-speculative execution. Under this model, a correct prediction incurs no cost, while an incorrect prediction can instead cause a slowdown; 
On clean inputs, we seek the speedup from correct speculation, so abstention is charged for forfeiting that speedup.

\paragraph{Oblivious vs.\ adaptive adversaries.} We also distinguish between two adversarial models. An \emph{oblivious} adversary must fix prior to the interaction its corruption rounds $(c_t)_{t=1}^T$ as well as every corrupted instance $(x_t)_{t:c_t=1}$ (potentially using randomness). 
An \emph{adaptive}
adversary may instead be adaptive to the history and depend on all previous learner decisions as well as clean samples on previous clean rounds. We remark that on round $t$ it must nevertheless commit to $c_t$ before observing a possible fresh clean draw $x_t\sim\mu$.

\begin{remark}
    \label{remark:exchangeable_setting}
    The guarantee proofs for our algorithms in the unknown-clean-distribution with oblivious adversaries (for full feedback) only rely on the exchangeability of clean samples. As such, our results in this setting directly apply to the following stronger model in which the adversary may select clean samples as well, but not their arrival order:

    Before the interaction begins, an adversary selects an unknown target $f^\star\in\Fcal$, the sequence of corruption rounds $(c_t)_{t\in[T]}$, all instances on corrupted rounds $(x_t)_{t:c_t=1}$, as well as the collection $S\subseteq \Xcal$ containing $C:=|\{t:c_t=0\}|$ instances. Clean samples $(x_t)_{t:c_t=0}$ are then generated as a uniformly random permutation of the instances in $S$. With the sequence $(x_t)_{t\in[T]}$ defined, the interaction proceeds as in the original learning protocol.
\end{remark}

\comment{

For the analysis, let $\mathcal H_{t^-}$ denote the information available immediately before the
adversary makes this round-$t$ commitment.  It contains the entire past but not $c_t$, the
current instance, or the learner's fresh round-$t$ randomization.  The timing convention ensures
that
\[
  \mathcal L(x_t\mid\mathcal H_{t^-},c_t=0)=\mu.
\]

\paragraph{Standing conventions.}
We assume that all shattering events and sample statistics used below are measurable, including the
subset averages and median estimates in the adaptive analysis.  Otherwise, probability and
expectation may be interpreted in the usual outer-probability sense.  This convention does not
affect the combinatorial arguments.

Our results are information-theoretic.  Each algorithm is assumed to have exact access to the
shattering probabilities, counts, or estimates it invokes, or to an oracle that returns them.  We
do not claim computationally efficient implementations.
}

\paragraph{Notations.} For $n\in\mathbb N$, write $[n]:=\{1,\ldots,n\}$.
We write $H_n:=\sum_{i=1}^n1/i$ for the $n$-th harmonic number. 
For any sequence $S=(z_1,\ldots,z_n)$ and $U\subseteq[n]$, we denote by $S_{-U}$ the sequence obtained by deleting coordinates in $U$. We also abbreviate $S_{-\{i\}}$ by $S_{-i}$. For a new element $z$, $S\Vert z$ denotes the sequence obtained from $S$ by appending $z$.
If $S$ is a sequence of labeled datapoints $z_i=(x_i,y_i)\in\Xcal\times\{0,1\}$, we write $S[i\gets y]$ for the sequence obtained by replacing the label at coordinate $i$ by $y$.

We next detail notations for operations on function classes. The algorithms may update their version space and restrict it after observing a label. Formally, for any
$V\subseteq\F$, $x\in\X$, and $y\in\{0,1\}$, we denote by
\[
  V^{x\to y}:=\{f\in V:f(x)=y\}
\]
the subclass of $V$ that assigns label $y$ to $x$.  

For a function $f$, a function class $H$, and a set $A$, let $f|_A$ denote the
restriction of $f$ to $A$, and let $H|_A:=\{h|_A:h\in H\}$ the projection of $H$ onto $A$. We write $V\shatters A$ to say that $V$
\emph{shatters} $A$.
\section{Known Clean Distribution}
\label{sec:known}

We begin with the setting in which the clean distribution $\mu$ is known and feedback is full. The adversary may still adapt to the observed past.  The natural quantities in this setting are the shattering probabilities: for a version space $V$, one asks for the probability that $V$ shatters $k$ fresh draws from $\mu$.

The learner of \citet{goel2023adversarial} follows one such probability at a time.  It starts at level $d$, and a mistake reduces the active shattering probability by a constant factor.  Once that probability becomes small, the learner moves to the next level.  With the threshold $\alpha_k=T^{-k}$, this takes $O(\log(1/\alpha_k))=O(k\log T)$ mistakes at level $k$; summing over all levels gives $O(d^2\log T)$.  The one-point argument at each level also gives $O(d)$ expected clean abstentions.  Thus the quadratic mistake bound comes from restarting the accounting whenever the level changes.

Our learner avoids these restarts.  It combines all shattering probabilities, with geometric weights, into one potential.  The weights are chosen so that a mistake contracts the potential and so that the contribution of extending a shattered $k$-tuple by one clean point can be charged to the term at level $k+1$.  Both losses can then be analyzed using the same quantity.

\begin{definition}[Shattering probabilities]
\label{def:population-shattering}
For $V\subseteq\F$, let $X_1,\ldots,X_k$ be independent draws from $\mu$ and define
\[
  \rho_k(V)
  :=\Pp_{X_1,\ldots,X_k\sim\mu}
      \bigl(V\shatters(X_1,\ldots,X_k)\bigr).
\]
Set $\rho_0(V):=\1\{V\neq\varnothing\}$.
\end{definition}

Thus $\rho_k(V)$ measures the ambiguity remaining in $V$ at sample size $k$.  It vanishes for
$k>d$.

\begin{definition}[Weighted shattering potential]
\label{def:population-potential}
For $\beta>0$, define
\[
  \Theta_\beta(V):=\sum_{k=0}^d\beta^k\rho_k(V).
\]
\end{definition}

The parameter $\beta$ controls the mistake--abstention tradeoff.  First, restricting $V$ can only
decrease the potential, and $\Theta_\beta(V)\geq1$ whenever $V\neq\varnothing$.  Since
$\rho_k(V)\leq1$, the potential lies in $[1,(1+\beta)^d]$ while the target remains in the version
space.  Consequently, a constant-factor contraction can occur only
$O(d\log(1+\beta))$ times.

Second, the geometric weights match the one-point decomposition in \cref{lem:one-point}.
Averaging the extension term at level $k$ produces $\rho_{k+1}(V)$, and
\[
  \sum_{k=0}^{d-1}\beta^k\rho_{k+1}(V)
  =\frac1\beta\sum_{j=1}^d\beta^j\rho_j(V)
  \leq\frac1\beta\Theta_\beta(V).
\]
This shift by one level is why the same potential controls clean abstentions without an additional
sum over $k$.

At round $t$, let $V_t$ be the hypotheses consistent with the labels revealed so far.  The learner
compares the two restricted potentials and uses the three-quarter threshold that makes every
mistake contract the potential; the complete rule is given in \cref{alg:known}.

\begin{algorithm}
\caption{Known-distribution abstaining learner}
\label{alg:known}
\begin{algorithmic}
\State \textbf{Require:} class $\F$, clean distribution $\mu$, and $\beta>0$.
\State Initialize $V_1\leftarrow\F$.
\State \textbf{For} $t=1,\ldots,T$:
\State \quad observe $x_t$ and set $P_b\leftarrow\Theta_\beta(V_t^{x_t\to b})$ for
                 $b\in\{0,1\}$;
\State \quad if $\min\{P_0,P_1\}>\tfrac34\Theta_\beta(V_t)$, output
                 $\widehat y_t\leftarrow\bot$;
\State \quad otherwise output
                 $\widehat y_t\leftarrow\argmax_{b\in\{0,1\}}P_b$, breaking ties deterministically;
\State\quad if $\widehat y_t\notin\{y_t,\bot\}$, observe $y_t$ and update $V\gets V^{x_t\to y_t}$ \Comment{update version space upon mistakes}
\end{algorithmic}
\end{algorithm}


\begin{theorem}[Known distribution]
\label{thm:known}
For every $\beta>0$ and every adaptive adversary, 
\cref{alg:known} satisfies
\[
  \MisErr\leq\frac{d\log(1+\beta)}{\log(4/3)}\qquad \text{and} \qquad
  \E[\AbsErr]\leq\frac{2T}{\beta}.
\]
The expectation is over the clean draws and any randomization of the adversary and learner.
\end{theorem}

The proof has two parts.  Every mistake follows the smaller branch and contracts the potential,
which proves the realized-interaction mistake bound.  An abstention requires both branches to be
large; the one-point decomposition shows that this is rare when the current point is a clean draw.

Taking $\beta=T$ gives $O(d\log T)$ mistakes on every realized interaction and at most two
expected clean abstentions.  This removes one factor of $d$ from each bound of
\citet{goel2023adversarial} and resolves their question.
More generally, choosing $\beta=T/\delta$ for any $\delta\in(0,1]$ gives
$O(d\log(T/\delta))$ mistakes and at most $2\delta$ expected clean abstentions.

\subsection{Mistakes contract the potential}

\begin{proposition}[Mistake bound for every realized interaction]
\label{prop:known-mistakes}
On every realized interaction,
\[
  \MisErr\leq\frac{d\log(1+\beta)}{\log(4/3)}.
\]
\end{proposition}

\begin{proof}
Set $\Phi_t:=\Theta_\beta(V_{t+1})$ for $t=0,\ldots,T$.  Since the target $f^\star$ remains in
the version space, $\Phi_t\geq1$.  Moreover,
\[
  \Phi_0=\Theta_\beta(\F)\leq\sum_{k=0}^d\beta^k\leq(1+\beta)^d.
\]
Monotonicity gives $\Phi_t\leq\Phi_{t-1}$ on every round.  If a mistake occurs, the update follows
the branch not selected by the learner.  Therefore,
\[
  \Phi_t=\min\{P_0,P_1\}\leq\tfrac34\Phi_{t-1}.
\]
If $M=\MisErr$, iterating these inequalities gives
\[
  1\leq\Phi_T\leq\left(\frac34\right)^M(1+\beta)^d.
\]
Taking logarithms proves the proposition.
\end{proof}

\subsection{A clean point rarely triggers abstention}

Potential contraction alone cannot bound abstentions, because an abstention need not change the
version space.  Instead, we show that both branches can be large only when the current point extends
unusually many shattered samples.  We first quantify this extension event.  For
$k\geq1$, define
\[
  \sigma_k(V,x)
  :=\Pp_{X_1,\ldots,X_k\sim\mu}
      \bigl(V\shatters(x,X_1,\ldots,X_k)\bigr).
\]
Set
$\sigma_0(V,x):=\1\{V^{x\to0}\neq\varnothing,\ V^{x\to1}\neq\varnothing\}$.
The tuple $(x,X_1,\ldots,X_d)$ cannot be shattered: either it contains a repeated point, or it
contains $d+1$ distinct points, which is impossible for a class of VC dimension $d$.  Hence
$\sigma_d(V,x)=0$.  Moreover, for
$0\leq k\leq d-1$,
\begin{equation}
  \E_{x\sim\mu}[\sigma_k(V,x)]=\rho_{k+1}(V).
  \label{eq:anchored-average}
\end{equation}
Thus, if the anchor $x$ is drawn independently from $\mu$, the mean probability of an anchored $k$-extension is exactly $\rho_{k+1}(V)$.

The next lemma is the weighted form of the fixed-level one-point inequality of \citet{goel2023adversarial}.

\begin{lemma}[One-point decomposition]
\label{lem:one-point}
For every $V\subseteq\F$ and $x\in\X$,
\[
  \Theta_\beta(V^{x\to0})+\Theta_\beta(V^{x\to1})
  \leq \Theta_\beta(V)+\sum_{k=0}^{d-1}\beta^k\sigma_k(V,x).
\]
\end{lemma}

\begin{proof}
For $A=(X_1,\ldots,X_k)$, pointwise
\[
  \1\{V^{x\to0}\shatters A\}+\1\{V^{x\to1}\shatters A\}
  \leq \1\{V\shatters A\}+\1\{V\shatters(x,A)\}.
\]
This is the fixed-level indicator inequality of \citet{goel2023adversarial}.  Taking expectations gives
\[
  \rho_k(V^{x\to0})+\rho_k(V^{x\to1})
  \leq\rho_k(V)+\sigma_k(V,x).
\]
Multiplying by $\beta^k$, summing over $k\leq d$, and using $\sigma_d=0$ proves the claim.
\end{proof}

If the learner abstains, the two branch potentials sum to more than $3\Theta_\beta(V)/2$, so the decomposition forces the anchored extension sum to exceed $\Theta_\beta(V)/2$.  For a clean $x$, the expected anchored extension sum is at most $\Theta_\beta(V)/\beta$.  Markov's inequality therefore gives the following conditional bound.

\begin{proposition}[Conditional abstention bound]
\label{prop:known-abstentions}
Conditional on the pre-round history and on round $t$ being clean,
\[
  \Pp(\widehat y_t=\bot\mid\mathcal H_{t^-},c_t=0)\leq\frac{2}{\beta}.
\]
\end{proposition}

\begin{proof}
Condition on the realized pre-draw history and write $V=V_t$.  On a clean round, $x_t\sim\mu$.  If the learner abstains, both restrictions have potential greater than $3\Theta_\beta(V)/4$.  Hence, \cref{lem:one-point} implies
\[
  \sum_{k=0}^{d-1}\beta^k\sigma_k(V,x_t)>\tfrac12\Theta_\beta(V).
\]
By \cref{eq:anchored-average},
\[
  \E_{x_t\sim\mu}\left[\sum_{k=0}^{d-1}\beta^k\sigma_k(V,x_t)\right]
  =\sum_{k=0}^{d-1}\beta^k\rho_{k+1}(V)
  =\frac1\beta\sum_{j=1}^d\beta^j\rho_j(V)
  \leq\frac1\beta\Theta_\beta(V).
\]
The result follows from Markov's inequality.
\end{proof}

\begin{proof}[Proof of \cref{thm:known}]
The mistake claim is \cref{prop:known-mistakes}.  On a corrupted round the charged abstention indicator is zero.  On a clean round, \cref{prop:known-abstentions} bounds its conditional expectation by $2/\beta$.  Summing over $t\in[T]$ proves the expectation bound.
\end{proof}

The next section replaces population probabilities by deterministic weighted counts.

\section{Unknown Distribution with Oblivious Corruptions}
\label{sec:unknown-oblivious}

When the clean distribution is unknown, the population shattering probabilities used above are
unavailable.  Ordinary empirical frequencies are not a valid substitute: the observed history
contains injected points and need not resemble an i.i.d.\ sample from the clean distribution.  We
make this failure concrete, then retain the underlying shattering witnesses as a deterministic
potential.  Against an oblivious adversary, the resulting algorithm has square-root dependence on
$T$, improving the previous three-quarter-power dependence
\citep{yu2026distributionfreev2}.

\subsection{A Weighted History Potential and Algorithm}
\label{sec:empirical-potential}

Let $S=(z_1,\ldots,z_n)$ be a finite labeled history, where
$z_j=(x_j,y_j)\in\X\times\{0,1\}$.  


\paragraph{Shattered subsets.}
The natural analogue of a fresh shattered sample is a set of coordinates whose labels remain free
after all other observed labels have been enforced.  A set $U\subseteq[n]$ is
\emph{shattered in $S$} if, for every $b\in\{0,1\}^U$, there is an $f\in\F$ such that
\[
  f(x_j)=y_j\quad(j\notin U),
  \qquad
  f(x_j)=b_j\quad(j\in U).
\]
Let $\Sh(S)$ be the family of such subsets. Membership $U\in\Sh(S)$ means that the observed labels remain enforced at every coordinate
outside $U$, while the labels on $U$ may vary arbitrarily. 

This definition suggests a direct empirical replacement for $\rho_k$.  For each $k$-subset $U$,
enforce the labels in $S_{-U}$ and ask whether the resulting version space shatters the held-out
coordinates in $U$.  Averaging this indicator gives the leave-$k$-out statistic
\[
  \widetilde\rho_k(S):=
  \frac{|\{U\in\Sh(S):|U|=k\}|}{\binom{n}{k}},
\]
the fraction of $k$-subsets of the observed history that are shattered in this sense.  If the
history were i.i.d.\ from $\mu$, this average would have the desired sampling interpretation.

Injected duplicates show why that interpretation fails here.  Suppose a labeled point has already
imposed its label constraint, and append $m$ identical injected copies.  The version space and its
population shattering behavior do not change.  However, none of the $m$ singleton deletions of
those copies is shattered, because another copy retains the same constraint.  Only the original
coordinates can contribute to the numerator of $\widetilde\rho_1(S)$, while its denominator grows
by $m$.  The adversary can therefore drive $\widetilde\rho_1(S)$ to zero without changing the
population quantity it is meant to represent.  More generally, normalized leave-out frequencies
need not reflect the shattering behavior of clean draws under $\mu$.

We therefore retain the combinatorial witnesses but discard the normalization and the probability
interpretation.

\begin{definition}[Weighted history potential]
\label{def:empirical-potential}
For $\beta>0$, define
\[
  \widehat\Theta_\beta(S):=
  \sum_{U\in\Sh(S)}\beta^{|U|}.
\]
\end{definition}

Each shattered subset contributes its deterministic weight regardless of how many injected points
appear in the history.  Thus $\widehat\Theta_\beta(S)$ records unresolved sets at every size; it is
not intended to estimate the population potential.

If $S$ is realizable, then $\varnothing\in\Sh(S)$.  Every shattered subset has size at most $d$,
because its points are shattered in the ordinary VC sense by hypotheses consistent outside that
subset.  Consequently, $\widehat\Theta_\beta(S)$ contains no term of degree greater than $d$.
It is also unchanged if the entries of $S$ are permuted, provided the subsets are reindexed by the
same permutation.  This permutation invariance is what will allow us to replay the clean examples
in a different order.

Write $S\Vert z$ for appending a labeled example $z$.  The learner uses the same three-quarter
contraction rule as in the known-distribution case, now with the weighted history potential.  For
each candidate label, it evaluates the potential after appending that label.  It predicts the label
of the larger branch only when the other branch is small enough; if the prediction is wrong, the
revealed label therefore sends the history down the smaller branch. 

Finally, for a more refined performance guarantee, we make the following slight adjustment to the weighted history potential. For any dataset $S=(x_u,y_u)_{u\leq t}$ we define the \emph{age-dependent} weight of datapoint $(x_u,y_u)$ via
\begin{equation*}
    \lambda_{u,t}:=\beta\paren{1+(t-u)\frac{\beta}{d}}.
\end{equation*}
Importantly, the older the datapoint, the larger weight it has. We then consider the following age-weighted potential, which corresponds to the partition function for the distribution of shattered sets in $S$ induced by the age-dependent weights.
\begin{definition}[Age-weighted history potential]
\label{def:weighted_partition_function}
    We define the age-weighted history potential as
    \begin{equation*}
        \widehat \Theta_{\lambda} (S):=\sum_{U\in\Sh(S)} \prod_{u\in U} \lambda_{u,t}.
    \end{equation*}
\end{definition}
The complete algorithm based on age-weighted history potential is detailed in \cref{alg:empirical}. As a remark, if we replace $\widehat \Theta_{\lambda} (S)$ by the simpler potential $\widehat \Theta_{\beta} (S)$ in \cref{alg:empirical}, we lose a logarithm factor in $T$ (see \cref{thm:unknown-oblivious}). In comparison, the age-dependent potential places more mass on older datapoints and since the learner aims to minimize that potential, this incentivizes the learner to prioritize resolving older ambiguities. For intuition, resolving older datapoints ensures that if the learner made abstentions, it was because the adversary significantly modified the distributions of observed instances---in turn, this prevents the abstention error from growing logarithmically due to old observations. Altogether, \cref{alg:empirical} enjoys the following guarantee.

\begin{algorithm}
\caption{Weighted-history abstaining learner}
\label{alg:empirical}
\begin{algorithmic}
\State \textbf{Require:} class $\F$ and parameter $\beta>0$.
\State Initialize $S_0\leftarrow()$ and $\lambda_{u,t}\leftarrow\beta\paren{1+(t-u)\frac{\beta}{d}}$ for all $1\leq u\leq t\leq T$.
\State \textbf{For} $t=1,\ldots,T$:
\State \quad observe $x_t$ and set
  $P_b\leftarrow\widehat\Theta_\lambda(S_{t-1}\Vert(x_t,b))$ for $b\in\{0,1\}$;
\State \quad if $\min\{P_0,P_1\}>\tfrac34\widehat\Theta_\lambda(S_{t-1})$, output
  $\widehat y_t\leftarrow\bot$;
\State \quad otherwise output
  $\widehat y_t\leftarrow\argmax_{b\in\{0,1\}}P_b$, breaking ties deterministically;
\State \quad observe $y_t$ and update $S_t\leftarrow S_{t-1}\Vert(x_t,y_t)$.
\end{algorithmic}
\end{algorithm}

\begin{theorem}[Unknown distribution, oblivious corruption]
\label{thm:unknown-oblivious-no-log} There is a universal constant $C>0$ such that for every $\beta\in(0,1/4]$, \cref{alg:empirical} satisfies, on every realizable labeled interaction,
\begin{equation}
  \MisErr
  \leq \frac{2\beta}{\log(4/3)+2\beta} T
  \label{eq:unknown-mistake-bound}
\end{equation}
Against an oblivious adversary,
\begin{equation}
  \E[\AbsErr]
  \leq  C\frac{d}{\beta} + Cd\log \paren{1+\frac{\beta T}{d}}.
  \label{eq:unknown-abstention-bound}
\end{equation}
The expectation is over the clean draws and any randomization of the adversary and learner.
\end{theorem}

The mistake rate in \eqref{eq:unknown-mistake-bound} is proportional to $\beta$, whereas for small $\beta$, \eqref{eq:unknown-abstention-bound} is of order $d/\beta$. 
\begin{corollary}[Oblivious full feedback rate]
\label{cor:unknown-balanced}
For every $\alpha\in[0,1)$ \cref{alg:empirical} achieves
  \[
    \MisErr=O(T^\alpha)
    \qquad\text{and}\qquad
    \E[\AbsErr]=O(dT^{1-\alpha}).
  \]
  In particular, when $T\geq 16d$, balancing misclassification and abstention error for $\beta=\sqrt{d/T}$ gives
  \begin{equation*}
      \MisErr + \E[\AbsErr] = O(\sqrt{dT}).
  \end{equation*}
\end{corollary}



\subsection{Potential Growth and Mistake Contraction}
\label{sec:unknown-proof-roadmap}
In this subsection we will provide some intuitions on how \cref{alg:empirical} achieves its performance guarantee. For simplicity, we will focus on \cref{alg:empirical} with $\widehat \Theta_{\lambda} (S)$ replaced by its non-age-dependent version $\widehat \Theta_{\beta} (S)$. Unlike the population potential for a known distribution, the history potential $\widehat \Theta_{\beta} (S)$ need not decrease when a labeled example
is appended, even on a non-mistake round.  The key replacement for monotonicity is that one append
operation increases it by at most a factor $1+\beta$.  Indeed, a shattered set in the extended
history either excludes the new coordinate, or includes it and becomes a shattered set of the old
history when that coordinate is removed.  Combining this growth bound with the algorithm's
$3/4$-contraction on mistakes and the lower bound $\widehat\Theta_\beta\geq1$ gives the
proposition below.

\begin{lemma}[One-step growth]
\label{lem:empirical-growth}
For every labeled history $S$ and every labeled example $z$,
\[
  \widehat\Theta_\beta(S\Vert z)
  \leq(1+\beta)\widehat\Theta_\beta(S).
\]
\end{lemma}

\begin{proposition}[History-potential mistake bound for every realizable interaction]
\label{prop:empirical-mistakes}
For every realizable interaction, \cref{alg:empirical} with non-age-dependent potential satisfies
\[
  \MisErr
  \leq
  \frac{\log(1+\beta)}{\log(4(1+\beta)/3)}\,T.
\]
\end{proposition}

The exact coefficient follows from the same accounting.  If $M$ mistakes occur, realizability and
the one-step bounds give
\[
  1
  \leq \widehat\Theta_\beta(S_T)
  \leq \left(\frac34\right)^M(1+\beta)^{T-M}.
\]
Taking logarithms yields \cref{prop:empirical-mistakes}.  The formal combinatorial proof of the
growth lemma appears in \cref{app:unknown-upper-proofs}.

\subsection{Clean Abstentions from Bad Positions and Exchangeability}

We now turn the deterministic count into a clean-round abstention bound.  The argument has three
steps.

\paragraph{Step 1: replay one coordinate.}
Fix a realizable history $S=(z_1,\ldots,z_n)$, delete coordinate $i$, and imagine presenting
$x_i$ after all the other examples.  We recall the notation $S[i\leftarrow y]$ for the sequence obtained by
replacing the $i$-th label with $y$.  If the replayed point receives candidate label $y$,
permutation invariance shows that the resulting branch potential is
\[
  P_i(y):=\widehat\Theta_\beta(S[i\leftarrow y]).
\]
Thus $P_i(0)$ and $P_i(1)$ are exactly the two values that the learner would compare in this
replay.  We call $i$ \emph{bad in $S$} if the learner would abstain:
\[
  \min\{P_i(0),P_i(1)\}>\tfrac34\widehat\Theta_\beta(S_{-i}),
\]
and write $\operatorname{Bad}(S)$ for the set of bad indices.

\paragraph{Step 2: only a few coordinates are bad.}
For each coordinate, define its local witness weight
\[
  W_i(S):=\sum_{\substack{U\in\Sh(S)\\i\in U}}\beta^{|U|-1}.
\]
This quantity records how much of the empirical potential can be charged to coordinate $i$. Each shattered set $U$ contributes once to $W_i(S)$ for every $i\in U$.  Consequently
\[
  \sum_{i=1}^n W_i(S)
  =\sum_{U\in\Sh(S)}|U|\beta^{|U|-1}
  \leq \frac d\beta\widehat\Theta_\beta(S),
\]
where the last inequality uses $|U|\leq d$. The next lemma relates the two branch potentials to the local witness weight $W_i (S)$.


\begin{lemma}[Replay-branch decomposition]
\label{lem:empirical-pointwise}
For every realizable history $S$ and every $i\in[n]$,
\[
  P_i(0)+P_i(1)
  \leq
  \widehat\Theta_\beta(S_{-i})+(1+2\beta)W_i(S).
\]
\end{lemma}
The proposition then sums those local weights to show that there are few bad positions.
\begin{proposition}[Few bad indices]
\label{prop:few-bad-indices}
For every realizable history $S$,
\[
  |\operatorname{Bad}(S)|
  \leq
  \frac{2d(1+2\beta)(1+\beta)}{\beta}.
\]
\end{proposition}

Indeed, the replay-branch decomposition and the one-step growth lemma force every bad coordinate
to carry a fixed fraction of $\widehat\Theta_\beta(S)$ in local witness weight.  The total-weight
bound above then gives \cref{prop:few-bad-indices}.  This deterministic counting argument is proved
in \cref{app:unknown-upper-proofs}.

\paragraph{Step 3: use exchangeability to identify the current clean coordinate.}
Condition on the oblivious adversary's private seed, which fixes the injected examples and their
locations.  At the $i$th clean round, permutation invariance lets us place the injected history
first and replay the first $i$ clean labeled examples afterward.  The learner's current decision
is then the event that the final clean coordinate is bad.  The $i$ clean coordinates are
exchangeable, so the final one is bad with probability at most the total number of bad coordinates
divided by $i$.
Combining this observation with \cref{prop:few-bad-indices} gives the next proposition; summing its
$1/i$ bound gives \eqref{eq:unknown-abstention-bound} up to a $\log T$ factor.  The formal replay and conditioning
argument appears in \cref{app:unknown-upper-proofs}.

\begin{proposition}[Clean-round exchangeability]
\label{prop:clean-round-exchangeability}
For the $i$th clean round $s_i$ of any fixed oblivious corruption schedule,
\[
  \Pp(\widehat y_{s_i}=\bot)
  \leq
  \frac{2d(1+2\beta)(1+\beta)}{\beta i}.
\]
\end{proposition}
Finally, in the small-$\beta$ regime which we are interested in, summing up all $i$ in the clean rounds gives $\Ebb[\AbsErr]=O(\frac{d}{\beta}\log T )$. This, together with the deterministic $O(\beta T)$ mistake bound in \cref{prop:empirical-mistakes}, provides the guarantee in \cref{thm:unknown-oblivious-no-log} up to a logarithm factor for the simpler history potential $\widehat\Theta_\beta$. 

The proof of \cref{thm:unknown-oblivious-no-log} for the age-weighted history potential $\widehat\Theta_\lambda$ requires an additional analysis to account for the fact that the weights are now dependent on the clean sample assignment order. This modification essentially improves the abstention probability of \cref{prop:clean-round-exchangeability} from $O(\frac{d}{\beta i})$ to $O(\frac{d}{i}+(\frac{d}{\beta i})^2)$, removing the extra $\log T$ factor. The complete analysis of the age-weighted potential is given in \cref{app:unknown-upper-proofs-age-weighted}.

We next show that the dependence on $d$ and $T$ in \cref{thm:unknown-oblivious-no-log} cannot be improved.

\subsection{Oblivious Lower Bound}
\label{sec:unknown-lower-bound}
The lower bound combines two simple ideas.  First, even a VC-dimension-one class can make a new
point risky to label.  Second, the learner cannot tell which part of the sequence comes from the
clean distribution.  We use the first idea to create many rounds on which prediction is dangerous,
and the second to make abstention expensive on an unknown fraction of those rounds.

For the first idea, consider hypotheses defined by paths in a rooted $b$-ary tree.  A hypothesis
labels the nodes on one root-to-leaf path by one and all other nodes by zero.  At a node on the
hidden path, exactly one of its $b$ children is positive.  Until that child has appeared, it is
uniform among the children that remain possible.  Therefore, when a previously unseen child first
appears and at least two children are still possible, any prediction has conditional mistake
probability at least $1/b$.

For the second idea, repeat this experiment at $L$ successive depths, producing one row of
examples at each depth.  Choose $J\in[L]$ uniformly and declare the distribution used in row
$J$ to be the clean distribution.  The other rows are generated in advance and inserted by an
oblivious adversary.  Every row has the same labeled distribution whether or not it is the clean
one, so the transcript does not reveal $J$.  If $N_\bot$ is the total number of abstentions,
the random choice of the clean row gives
$\E[\AbsErr]=\E[N_\bot]/L$.

This hidden-clean-block argument is due to \citet{edelman2026reliable};
\citet{yu2026distributionfreev2} give a related tree-path lower bound using a recursive adversary.
To obtain the dependence on $d$ for an abstention budget $A$, set
$m:=\min\{d,A\}$ and take $m$ independent copies of the tree class, using the
same hidden row $J$ in every active copy.  We then pad the class with $d-m$ independent VC
coordinates that never appear in the process, so the resulting class has VC dimension exactly $d$.
Each row contains $\Theta(mb)$ risky first appearances in expectation, while only a $1/L$
fraction of all abstentions are charged.  The theorem below formalizes the resulting tradeoff.

\begin{theorem}[Oblivious lower bound]
\label{thm:unknown-lower-bound}
Fix $d\geq1,T\geq 32$ and any integer abstention budget $A$ satisfying
$1\leq A\leq T/32$. There is a class of VC dimension $d$ such that any full-feedback learner must suffer the following against some oblivious adversary:
\[
  \E[\AbsErr]>A
  \qquad\text{or}\qquad
  \E[\MisErr]\geq
  \frac1{1024}\min\left\{\frac{dT}{A},\,T\right\}.
\]
Consequently, for every $1\leq d\leq T/1024$, there is a class of VC dimension exactly $d$
such that every full-feedback learner must suffer the following against some oblivious adversary:
\[
  \E[\MisErr+\AbsErr]\geq \frac1{1024}\sqrt{dT}.
\]
\end{theorem}

Thus the budget tradeoff covers every $d$, with the mistake lower bound saturating at order $T$
when $d>A$.  For the balanced conclusion, take $A=\lfloor\sqrt{dT}\rfloor$; the condition
$d\leq T/1024$ ensures $1\leq A\leq T/32$, and this choice is within a
universal constant factor of $\sqrt{dT}$.  Together with the upper bound
$O(\sqrt{dT})$ from \cref{cor:unknown-balanced}, this gives the matching square-root rate in the
stated range.

\paragraph{Construction.}

Set $m:=\min\{d,A\}$.  Since $d$ and $A$ are positive integers,
$1\leq m\leq d$.
In each active copy $r\in[m]$, draw a path
$v_{r,0},v_{r,1},\ldots,v_{r,L}$ by choosing each next child uniformly.  For each depth $j$, let
the support of $P_j$ be the union, over $r\in[m]$, of the $b$ children of
$v_{r,j-1}$, and let $P_j$ be uniform on these $mb$ nodes.  Row $j$ consists of $mb$
independent draws from $P_j$.  The unique positive child in copy $r$ is $v_{r,j}$.
The active direct sum has VC dimension $m$.  Taking its product with all labelings of
$d-m$ additional points, which never appear in the process, gives VC dimension exactly $d$.

One row $J$ is chosen uniformly and its row distribution is declared to be the fixed clean
distribution.  Every draw in that row is clean, while all other rows are generated in advance as
injections from their own row distributions.  The labeled law of each row is independent of $J$;
only the unobserved clean/corrupt indicators change.

The shared row is important for the accounting.  If $N_\bot$ denotes the total number of
abstentions in all $L$ rows, indistinguishability and the uniform choice of $J$ give
\[
  \E[\AbsErr]=\frac1L\E[N_\bot].
\]
Call a first appearance \emph{uncertain} if it occurs before the positive child has been seen and
while at least two children remain possible.  To count these appearances, first expose the number
$N_r$ of a row's $mb$ draws that fall in active copy $r$.  Then
$N_r\sim\operatorname{Bin}(mb,1/m)$, so it has mean $b$ and, for $b\geq8$, is at least
$\lfloor b/2\rfloor$ with constant probability.  Conditional on that event, the one-copy coupon
calculation of \citet{edelman2026reliable} gives $\Omega(b)$ uncertain appearances in
copy $r$.  Summing over the $m$ active copies shows that a row contains $\Omega(mb)$
uncertain appearances in expectation.  Conditional on the learner's information after it observes
the current child, predicting on any such appearance has mistake probability at least $1/b$.
Consequently,
\[
  \E[\MisErr]
  \geq \frac1b\left(\frac{Lmb}{8}-\E[N_\bot]\right).
\]
If $\E[\AbsErr]\leq A$, the preceding identity gives
$\E[N_\bot]\leq AL$.  Choose
\[
  b:=\left\lceil\frac{16A}{m}\right\rceil,
  \qquad
  L:=\left\lfloor\frac{T}{mb}\right\rfloor.
\]
Because $m\leq A$ and $A\leq T/32$,
\[
  b\geq16,\qquad b\leq\frac{32A}{m},\qquad mb\leq T,
  \qquad L\geq\frac{T}{2mb}.
\]
The preceding inequality then gives
\[
  \E[\MisErr]\geq\frac{mT}{1024A}
  =\frac1{1024}\min\left\{\frac{dT}{A},\,T\right\}.
\]
The shared index is essential: it defines the single fixed distribution $P_J$, while the same
averaging identity charges abstentions across all active copies.

The complete direct-sum argument, parameter choices, treatment of randomized learners, and padding
of unused rounds appear in \cref{app:unknown-lower-proof}.

\section{Handling Censored Feedback for Oblivious Adversaries}
\label{sec:age-weighted-unlabeled-history}

In the previous section, we provided optimal algorithms against oblivious adversaries when the clean distribution $\mu$ is unknown. These are however designed for the full-feedback model in which a learner still observes a label on abstention rounds. In this section, we extend this methodology to the censored model, where the label is revealed only when the learner makes a prediction.

A tempting direct approach is to store every label that happens to be revealed
and run the age-weighted history learner on that labeled subsequence. This
fails because labels are then retained precisely on rounds when the learner
chooses to predict. The resulting history is selected according to the
learner's decisions, so the exchangeability argument from the preceding
section no longer applies.

We instead retain every observed instance without its label, forming an unlabeled history $U=(x_1,\ldots,x_t)$. Revealed labels
are used only after mistakes, when they restrict the current version space which will be denoted by $V_t$.
Thus censorship does not select which instances enter the history; it only selects a sequence of truthful version-space restrictions.
Similar to the previous history potentials---\cref{def:empirical-potential,def:weighted_partition_function}---we then use as potential the partition function of subsets of $U$ that are shattered by the current version space $V$, as defined below.

\begin{definition}[Unlabeled-history potential]
\label{def:age-weighted-unlabeled-potential}
For any version space $V\subseteq\Fcal$ and unlabeled history $U=(x_1,\ldots,x_t)$, define the age-weighted unlabeled-history potential via
\begin{equation}
    \widehat\Theta^u_t(V;U)
    :=
    \sum_{I\subseteq[t]}
    \mathbf 1\{V\triangleright U_I\}
    \prod_{i\in I}\lambda_{i,t}.
    \label{eq:age-weighted-unlabeled-potential}
\end{equation}
For the empty history, set
$\widehat\Theta^u_0(V;())=\mathbf 1\{V\neq\emptyset\}$.
\end{definition}
The potential takes a sum over the subsets of the observed instances that remain
shattered by the current version space. In particular, every contributing subset has size at most $d$. 
The learner uses the potential to make predictions or abstain as in the previous algorithms. However, the learner only updates the version space after each mistake, and ignores labels revealed after correct predictions.
It needs no label after an abstention, while every mistake is observed because every prediction is followed by its true label. The resulting algorithm is therefore implementable under censored feedback. 

For uniform concentration purposes, we periodically restart the version space $V\gets\Fcal$ every
\begin{equation}
 W_\beta:=\left\lceil\frac{d\log(T+1)}{\beta}\right\rceil,
 \label{eq:ar-block-length}
\end{equation}
iterations, forming epochs of length $W_\beta$. This in turn ensures that the potential number of version spaces arising throughout the procedure can be controlled. The algorithm is summarized in \cref{alg:censored-age-restarts} and has the following guarantee.

\begin{algorithm}[tbp]
 \caption{Censored learner with age-dependent weights and periodic restarts}
 \label{alg:censored-age-restarts}
 \begin{algorithmic}[1]
  \Require Class $\mathcal F$,
           parameter $\beta\in(0,1]$
  \State $U_0\gets()$ and
         $W_\beta\gets\lceil d\log(T+1)/\beta\rceil$
  \State\textbf{For} $t=1,\ldots,T$:
   \State\quad\textbf{If} $t\equiv1\pmod{W_\beta}$, $V\gets\mathcal F$ \Comment{Restart version space at beginning of epoch}
   \State\quad Observe $x_t$ and set $P_b\gets\widehat\Theta^u_t(V^{x_t\to b};U_{t-1})$ for $b\in\{0,1\}$
   \State \quad\textbf{If} $\min\{P_0,P_1\}>\frac34 \widehat\Theta^u_t(V;U_{t-1})$, output $\widehat y_t\gets\bot$
   \State\quad\textbf{Else} 
   \State\quad\quad output $\widehat y_t\gets\operatorname*{arg\,max}_{b\in\{0,1\}}P_b$,
           breaking ties deterministically, and observe $y_t$
    \State\quad\quad if $\widehat y_t\ne y_t$, update $V\gets V^{x_t\to y_t}$ \Comment{update only after a mistake}
   \State \quad update $U_t\gets U_{t-1}\mathbin\Vert x_t$
 \end{algorithmic}
\end{algorithm}

\begin{theorem}[Censored feedback with oblivious adversaries]
 \label{thm:censored-age-restarts}
 There is a universal constant $C>0$ such that, for every
$\beta\in(0,1]$,
\cref{alg:censored-age-restarts} satisfies, on every realizable
interaction,
\begin{equation}
  \label{eq:ar-mistakes}
    \operatorname{MisErr}\le\frac{4\beta T}{\log(4/3)}.
 \end{equation}
Also, against every oblivious adversary,
\begin{equation*}
  \mathbb E[\operatorname{AbsErr}]
  \le C\paren{\frac{d}{\beta}+d\log^3(2T)}.
 \end{equation*}
The expectation is over the clean draws and any randomization of the
oblivious adversary and learner.
\end{theorem}

Note that up to the additive abstention error $O(d\log^3(2T))$, this censored algorithm achieves the same guarantee as the full feedback algorithm \cref{alg:empirical} up to constant factors, and hence is constant-factor optimal.

The proof has the same two deterministic ingredients as the unknown-distribution proof. First, provided that the version space $V$ was not restarted on a given round $t$, the one-step growth argument of \cref{lem:empirical-growth} applies after labels are removed from the history. Conversely, each mistake decreases the potential by a factor of $3/4$ (provided the version space was not restarted). Additionally, using Sauer's lemma, each restart may increase the number of mistakes by at most $O(d\log T)$. This contributes to an additional $O(d\log T \cdot T/W_\beta)=O(\beta T)$ number of mistakes, which does not affect the order of magnitude of the misclassification error.

Second, the replay-branch decomposition is essentially the same as \cref{lem:empirical-pointwise}. 
The main new issue is that the version space is selected by the previous mistakes and may therefore depend on the clean ordering. We therefore prove a uniform version of the replay argument \cref{prop:clean-round-exchangeability} for each potential version space $V$. Once the version space is fixed, we can use a similar exchangeability argument as in the non-censored case: conditionally on the oblivious adversary's seed, and the unordered pool of clean samples, the samples observed on each clean round are exchangeable and can be permuted. An analog argument to the replay argument from \cref{prop:clean-round-exchangeability} essentially applies to each such fixed version space.

The version spaces arising throughout the procedure depend on the clean ordering. We therefore use a uniform concentration argument to bound the abstention probability simultaneously over every potential version space $V$.
Specifically, version spaces arising throughout the procedure correspond to constraints from misclassified datapoints: for $m$ such constraints, the number of version spaces is of order $T^m$. If we did not use restarting, $m$ would be at most the full misclassification error $O(\beta T)$ over all $T$ rounds. However, since we restart version spaces at the beginning of each epoch, the number of constraints $m$ is at most the misclassification on a single epoch, which can be shown to be $m=O(\beta W_\beta)=O(d\log T)$. Taking a union bound over all these version spaces adds a term depending only on the logarithm of their number: $m\log T$. This is the term that contributes to the additive $O(d\log^3 T)$ abstention error in \cref{thm:censored-age-restarts}. The formal argument appears in Appendix~\ref{app:censored-age-restarts}.

Taking $\beta=\sqrt{d/T}$ gives the following consequence.

\begin{corollary}[Balanced oblivious censored rate]
\label{cor:censored-unlabeled}
There is a universal consant $C>0$ and a censored-feedback learner such that, against every oblivious adversary for $T\geq d\log^6 d$,
\[
\operatorname{MisErr}\leq C\sqrt{dT} \quad \text{and} \quad 
\mathbb E[\operatorname{AbsErr}]
\leq C\sqrt{dT}.
\]
\end{corollary}

Only the replay step uses obliviousness. Conditional on an oblivious adversary's seed, the injected prefix is unchanged when the clean examples are permuted. Under adaptive injections, changing the order of earlier clean examples may change later injected instances. The candidate current clean examples are then not replayed inside one common history, so the bad-position count cannot be divided by the number of remaining clean coordinates.

\section{Learning Against Adaptive Adversaries}
\label{sec:adaptive-compression}

In this section, we handle the more challenging case of adaptive adversaries under both full and censored feedback. Recall that the weighted-history learner in \cref{sec:unknown-oblivious} fixes the injected sequence and replays the clean examples in a random order. Against an adaptive adversary, later injected examples may depend on earlier clean draws, so replaying those draws also changes the injected history. The adaptive proof must therefore isolate the i.i.d. clean sequence rather than replay
it through the contaminated history.

We build on the adaptive full-feedback reduction introduced by
\citet{yu2026distributionfreev2}.  Their reduction has three parts.  First, it analyzes an expert whose stored examples are all clean; these examples form its \emph{clean bank}. In their order of appearance, the bank examples are i.i.d., and every later clean example is a fresh draw from the
same distribution. Second, it uses reduction dimension, which bounds the number of constrained subclasses that can appear on a fixed sample, to obtain one guarantee for every version space the expert may reach. Third, because the learner does not know which rounds are clean, it aggregates
over candidate banks and compares itself with the candidate whose bank is actually clean. We improve these three parts using a normalized-mean expert, a leave-one-out bound together with a uniform concentration argument to compress the expert, and direct multiplicative-weights aggregation.

For comparison, their adaptive full-feedback theorem gives, for $\alpha\in[0,1/3]$,
\[
  \MisErr
  \lesssim T^{3\alpha},
  \qquad
  \E[\AbsErr]
  \lesssim
  d^2(D\log D+\log T)^{2/3}\log(T)\,T^{1-\alpha},
\]
where $D$ is their reduction dimension at order $d^2\log T$ constraints.
We obtain the following result.
\begin{theorem}[Adaptive adversaries via direct aggregation]
\label{thm:adaptive-full-feedback-direct}
There is a universal constant $C>0$ such that for every $T\geq 1$, $\epsilon\in(0,1]$, the following holds. Let
$\ell_T:=\lceil d\log(T+1)/\log(16/15)\rceil+1$ and suppose that $\mathcal F$ has
$\ell_T$-reduction dimension $D_T<\infty$.
Then there is a learner that satisfies against any adaptive adversary with censored feedback, without knowledge of $\mu$:
\[
  \MisErr\le C\frac{d^2D_T\log^4(2T)}{\epsilon},
  \qquad
  \E[\AbsErr]
  \le \epsilon T + CdD_T\log^2(2T).
\]
The mistake bound holds on every realized interaction.
\end{theorem}

In particular, this gives a tradeoff for any $\alpha\in[0,1]$:
\begin{equation*}
    \MisErr\lesssim T^\alpha + d^2D_T\log^4(2T),
  \qquad
  \E[\AbsErr]
  \lesssim d^2D_T T^{1-\alpha}\log^4(2T).
\end{equation*}
This matches the tradeoff from the oblivious case up to factors in the dimensions in $d,D_T$, and polylogarithmic factors. In particular, the dependency of the tradeoff in the horizon $T$ is tight up to logarithmic factors.

In the following, we give the main ingredients to prove this result. We start by constructing an algorithm that reaches the desired guarantees for the regime $\alpha\leq 1/2$ in \cref{sec:ac-scores,sec:adaptive-candidate-aggregation}, which respectively develop the crucial ingredients. We complete the regime for $\alpha\geq 1/2$ as well in \cref{subsec:completing_tradeoff} using a recursive version of the previous algorithm. 
Complete proofs appear in Appendix~\ref{sec:ac-proofs}.

\subsection{A Compressed Shattering-Witness Expert}
\label{sec:ac-scores}

We start by constructing a clean-bank expert whose bank consists of true clean samples, in the full-feedback model. Of course, the learner cannot directly identify this expert since it does not observe which rounds are clean; instead the learner considers a family of experts corresponding to different choices of bank rounds. In the aggregation step we will later show it can achieve similar performance to whichever expert matches the realized clean sequence.

\paragraph{Construction of clean bank scores.}
The clean bank will be used to estimate the weighted potential as follows. For a given block $S=(x_1,\ldots,x_N)\in\Xcal^N$ of size $N\geq d$, and any class $V\subseteq\Fcal$, we define the symmetric shattered-subset statistic
\[
  \widehat\rho_k^S(V)
  :=
  \binom Nk^{-1}
  \sum_{\substack{I\subseteq[N]\\|I|=k}}
  \mathbf 1\{V\shatters(x_i)_{i\in I}\},
  \qquad
  \widehat\rho_0^S(V):=\mathbf 1\{V\ne\varnothing\}.
\]
As in the weighted-potential learner from the known-distribution case, we combine the statistics at different levels into the single
block potential
\[
  \widehat\Theta_\beta^S(V)
  :=
  \sum_{k=0}^d\beta^k\widehat\rho_k^S(V).
\]
Following the analysis of the known-distribution case, a good choice is to take the parameter $\beta=T$. We use this choice from now on: for conciseness we write $\Theta^S:=\widehat\Theta_T^S$, dropping the subscript $\beta$.

\comment{
For every nonempty $W$,
\begin{equation}
 1\leq\Theta^S(W)\leq\sum_{k=0}^dT^k\leq(1+T)^d.
 \label{eq:ac-potential-range}
\end{equation}
}

For any considered class $V$, the expert estimates the weighted potential of version spaces $V^{x\to b}=\{f\in V: f(x)=b\}$
through a normalized combination of block estimates as defined below. Using multiple blocks significantly reduces the estimation failure probability, in the same manner as the median of block estimates from \cite{yu2026distributionfree}. Given bank blocks $S_1,\ldots,S_m$, we define the \emph{score}
\begin{equation}
 a_b(x;V):=\frac1m\sum_{j=1}^m
        \frac{\Theta^{S_j}(V^{x\to b})}{\Theta^{S_j}(V)},
 \qquad b\in\{0,1\}.
 \label{eq:ac-normalized-score}
\end{equation}
Notably, the normalization is performed separately within each block: The expert aims to predict when the ratio of potentials between the prior class $V$ and updated version space $V^{x\to b}$ is bounded away from one; the score $a_b(x)$ is the mean of these ratios computed on each block. 

\comment{
Although these scores are arithmetic means, mistakes will be
charged to the geometric-mean potential
$P(V):=(\prod_j\Theta^{S_j}(V))^{1/m}$.
}

\paragraph{A short description of scores.}
With this construction of scores, one requires $Nm$ clean samples: we may partition these into $m$ equal blocks of size $N$ each. Moreover, using larger blocks of size $N$ is necessary to improve the accuracy of the estimated weighted potentials. If we directly use the scores defined above, the learner therefore has to consider $T^{O(Nm)}$ experts, corresponding to the possible choices of $m$ blocks. 
However, the learner also aims to use as few experts as possible---within the aggregation strategy, the performance compared to the clean-bank expert will degrade with the logarithm of the number of experts.

To this end, we compress the description of scores using a uniform-concentration argument as follows.
Choose a block $j$ uniformly, then a subset $I\subseteq[N]$, $|I|\leq d$, with probability
\begin{equation*}
 \frac{T^{|I|}\ind\{V\triangleright S_{j,I}\}}
      {\binom N{|I|}\Theta^{S_j}(V)}.
\end{equation*}
Let $\nu_V$ denote the resulting distribution of the witness
tuple $Z=S_{j,I}$. Crucially, since a tuple shattered by $V^{x\to b}$ is
also shattered by $V$, we have
\begin{equation*}
 a_b(x;V)=\mathbb E_{Z\sim\nu_V}
              \1[V^{x\to b}\triangleright Z].
 \label{eq:ac-mixture-identity}
\end{equation*}

The compression argument relies on estimating the score $a_b(x)$ by an empirical mean of samples $Z\sim\nu_V$. The reduction dimension is then used to achieve uniform concentration of these empirical means. 
Formally, let 
\begin{equation*}
    \mathcal R_l:=\set{\Fcal\cap C: C\subseteq\Xcal\times\{0,1\}, |C|\leq l},\quad l\geq 1.
\end{equation*}
be the family of subclasses of $\mathcal F$ defined by at most $l$ labeled constraints. The VC dimension of
\[
 \left\{Z\mapsto\1[V\triangleright Z]:
                    V\in\mathcal R_{\ell_T}\right\}
\]
on tuples of length at most $d$ is precisely bounded by $O(D_T\log(2dD_T))$, where $D_T$ is the $\ell_T$-reduction dimension of $\Fcal$ as in \cref{thm:adaptive-full-feedback-direct}.
Indeed, the shattering indicators on $r$ such tuples are
determined by the trace of the function classes $V\in\Rcal_{\ell}$ on at most $rd$ instances, hence their number is bounded by $(rd)^{D_T}$.
Uniform convergence at a constant accuracy therefore ensures that there is a list of
\comment{\begin{equation}
 s:=\lceil C_sD_T\log(2dD_T)\rceil,
 \qquad \eta:=1/64,\qquad\theta:=29/32,
 \label{eq:ac-compression-parameters}
\end{equation}}
\begin{equation*}
    s:=\lceil C_sD_T\log(2dD_T)\rceil
\end{equation*}
witnesses $Z_1,\ldots,Z_s$ for some universal constant $C_s>0$, satisfying
\begin{equation}
 \sup_{x,b}\abs{\widehat a_b(x;V)-a_b(x;V)}\leq \frac{1}{64},\quad\text{where} \quad \widehat a_b(x;V):=\frac1s\sum_{u=1}^s
                \1[V^{x\to b}\triangleright Z_u].
 \label{eq:ac-uniform-compression}
\end{equation}
whenever $V$ is defined by at most $\ell_T-1$ constraints (see \cref{lem:ac-compression} for a proof).
Importantly, each witness uses at most $d$ bank instances; hence the total number of bank instances required to compute the approximate score $\widehat a_b(x)$ is at most $ds$, independent of $N$ and $m$.

\paragraph{Predictions and mistake updates.}
Similarly as in the previous algorithms, the expert abstains if both approximate scores exceed $\frac{3}{4}$, using valid lists of witnesses satisfying \cref{eq:ac-uniform-compression}; otherwise it predicts a label of maximum approximate score.
It restricts the version space $V$ to the true-label branch $V^{x_t\to y_t}$ only when it makes a mistake. Whenever the version space $V$ is updated, a new witness list $Z_1,\ldots,Z_s$ is used---this is required because the distribution over witnesses $\nu_V$ is dependent on the version space $V$. 

We can show that each mistake contracts the geometric mean of the block potentials, $P(V):=(\prod_{j=1}^m \Theta^{S_j}(V))^{1/m}$, by a factor at most
$3/4+1/64< 15/16$. On the other hand, using similar arguments as in the known-distribution case, weighted potentials for realizable version spaces lie within $[1,(1+T)^d]$. In turn, the number of mistakes with one fixed bank is then at most
\begin{equation*}
 M_T=\left\lceil d\log(1+T)/\log(16/15)\right\rceil=\ell_T-1,
 \label{eq:ac-mistakes-per-stage}
\end{equation*}
and hence version spaces arising during the procedure are defined by at most $\ell_T-1$ constraints as required for the approximate scores.

Meanwhile, an abstention forces more than $m/6$ blocks to have, roughly speaking, both branch potentials greater than $5/8$ of their current block potential. These calculations appear in Lemma~\ref{lem:ac-margins}. Using a replay argument, we can then show that for a fixed clean bank of size $Nm$, the abstention error rate is essentially $O(d/N)$ for each new clean round, as proved in \cref{lem:ac-uniform-certificate}.

We then consider the reference expert which abstains until $Nm$ clean rounds have passed; starting from the next round $t_0$ it then uses the clean bank for the sequence of witness lists $(\mathcal Z_r)_{0\leq r<\ell_T}$ which it uses to implement the strategy described above. We summarize the corresponding expert in \cref{alg:reference_process_no_restart}.
Altogether, we can show the following misclassification and abstention error bounds on this reference process. 

\begin{algorithm}[tbp]
\caption{Reference expert}
\label{alg:reference_process_no_restart}
\begin{algorithmic}[1]
\Require Class $\mathcal F$, start time $t_0$, witness-index lists $(\mathcal Z_r)_{0\leq r\leq M_T}$
\State Abstain $\widehat y_t\gets \bot$ for all $t<t_0$
\State Initialize $V\leftarrow\mathcal F$, $r\leftarrow0$
\State \textbf{For} each round $t\geq t_0$:
\State\quad observe $x_t$ and compute $\widehat a_b(x_t;V)$ for $b\in\{0,1\}$ using witness list $\mathcal Z_r$
\State\quad if $\min_{b\in\{0,1\}} \widehat a_b(x_t)>\frac{3}{4}$, abstain $\widehat y_t\gets \bot$
\State\quad otherwise, output $\widehat y_t\gets \arg\max_{b\in\{0,1\}}\widehat a_b(x_t)$, breaking ties deterministically
\State\quad observe $y_t$
\State\quad if $\widehat y_t\in\{0,1\}$ and $\widehat y_t\neq y_t$, update $V\gets V^{x_t\to y_t}$ and $r\gets r+1$ 
\Comment{update version space only after a mistake}
\end{algorithmic}
\end{algorithm}

\begin{proposition}\label{prop:performance_reference_expert_no_restart}
    There is a universal constant $C_m>0$ such that the following holds. Let $N\geq d$ and $\delta\in(0,1)$ and suppose that $\Fcal$ has $\ell_T$-reduction dimension $D_T$. If $m\geq C_m(D_T\log(2TD_T)+\log(2/\delta))$, then the reference expert has at most $M_T$ mistakes on every realized interaction and at most
    \begin{equation*}
        \frac{8dT}{N+1}+Nm+32+\delta T
    \end{equation*}
    expected abstention error.
\end{proposition}

This result contains the only probabilistic part of the reduction. It remains to describe how the learner can achieve comparable performance to the reference expert. Since it does not know which are the correct start time $t_0$ and witness-index lists $(\mathcal Z_r)_{0\leq r\leq M_T}$ we consider all experts arising from any possible choice for these quantities---Whenever a candidate expert’s prescribed action is undefined or cannot be evaluated using the information available at the current round, we let it take an arbitrary fixed action, such as abstaining; this convention does not affect the valid reference expert or its guarantees. Let $\Ccal$ be the resulting family: their number is $L:=|\Ccal|\leq (T+1)^{ds(M_T+1)+1}$.

\begin{remark}
    Using a recursive strategy, we will further remove the $Nm$ abstention error term from \cref{prop:performance_reference_expert_no_restart}, in \cref{subsec:completing_tradeoff}. This improvement is not significant if the first error term $O(dT/N)$ dominates the abstention error bound---this precisely corresponds to the regime where the abstention error is $T^{\alpha}$ for $\alpha>1/2$.
\end{remark}

\subsection{Direct Aggregation over Candidate Experts}
\label{sec:adaptive-candidate-aggregation}

Yu and Blanchard handle this final step with their layered \textsc{Boosting} procedure.  If one
expert has at most $M$ mistakes, their guarantee adds $O(\varepsilon T)$ abstentions at the
cost of $O((M+\log T)(\log L)^2/\varepsilon)$ mistakes, where $L$ is the number of
candidate-bank experts.  Using a single weight potential, we give a sharper
comparison: it adds the same $O(\varepsilon T)$ abstentions with only
$O((M+\log L)/\varepsilon)$ mistakes.  This is our replacement for the final step of their
reduction.

The aggregate assigns a weight to each of the $L$ experts.
On each round, the score $P_b$ is the total weight of experts that
do not oppose label $b$: those that predict $b$ or abstain. The learner abstains precisely
when less than a $\kappa$ fraction of the total weight predicts each label, then rewards the experts that did predict by multiplying their weight by $1+\gamma$. Otherwise, the learner predicts the branch with larger weight, and halves the weights of experts that were found making a mistake. We state the complete rule in
\cref{alg:adaptive-direct-aggregation} and reprove the specialized guarantee in the appendix so
that the argument is self-contained. As a remark, this direct aggregation is fully implementable in the censored model provided access to the experts.

\begin{algorithm}
\caption{Direct aggregation of candidate-bank experts}
\label{alg:adaptive-direct-aggregation}
\begin{algorithmic}
\State \textbf{Require:} expert family $\mathcal C$, margin $\kappa\in(0,1/2)$, and update
  rate $\gamma\in(0,1]$.
\State Initialize $w_{c,1}\gets1/|\mathcal C|$ for every $c\in\mathcal C$.
\State\textbf{For} $t=1,\ldots,T$:
\State \quad observe $x_t$, obtain each expert's recommendation
  $\widehat y_{c,t}\in\{0,1,\bot\}$, and set
  \[
    \Phi_t\gets\sum_{c\in\mathcal C }w_{c,t},
    \qquad
    P_b\gets
    \sum_{c\in\mathcal C }
    w_{c,t}\mathbf 1\{\widehat y_{c,t}\in\{b,\bot\}\},
    \quad b\in\{0,1\}
  \]
\State \quad \textbf{If} $\min\{P_0,P_1\}>(1-\kappa)\Phi_t$ \text{then}
\State \quad\quad output
  $\widehat y_t\gets\bot$ and update $w_{c,t+1}\gets w_{c,t}(1+\gamma)^{\1[\hat y_{c,t}\neq\bot]}$ for each $c\in\mathcal C $
\State\quad \textbf{Else} 
\State\quad\quad output
  $\widehat y_t\gets\argmax_{b\in\{0,1\}}P_b$, breaking ties deterministically
  \State \quad\quad observe $y_t$ and update $w_{c,t+1}\gets w_{c,t}2^{-\1[\hat y_{c,t}= 1-y_t]}$ for each $c\in\mathcal C $
\end{algorithmic}
\end{algorithm}

An abstaining expert contributes to both $P_0$ and $P_1$.  Consequently, a learner mistake
forces a fixed fraction of the total weight to lie on incorrect experts and contracts the total
weight.  Conversely, whenever the learner abstains while the matching clean-bank expert predicts,
the update increases that expert's relative weight.  
Comparing total weight with the weight of one expert gives
the following pathwise guarantee.

\begin{proposition}[Direct aggregation]
\label{prop:adaptive-direct-aggregation-guarantee}
Let $M\geq 0$ and $R:=1+\log L+(\log2)M$.  Fix
$\kappa\in(0,1/2)$, suppose $R\le2\kappa T$, and that there is an expert which makes at most $M$ mistakes. Run
\cref{alg:adaptive-direct-aggregation} with
$\gamma=R/(2\kappa T)$. Then, on every realized interaction,
\[
  \MisErr
  \le
  \frac{4R}{\kappa}.
\]
Moreover, simultaneously for every expert with at most $M$ mistakes and every
$U\subseteq[T]$, the learner abstains on at most $8\kappa T$ more rounds of $U$ than that
expert.
\end{proposition}

The pathwise weight comparison proving this proposition appears in
\cref{app:adaptive-direct-aggregation}.

Applying the proposition to that set transfers the reference
expert's expected clean-abstention bound to the learner, with an additional $8\kappa T$ term.
This aggregation step is entirely pathwise; adaptivity enters only through the matching
clean-bank guarantee above.

For a target clean-abstention rate $\varepsilon$, we choose the parameters $\kappa=\varepsilon/8$, which limits the aggregation overhead to $O(\varepsilon T)$ abstentions. We next choose $\delta=1/T$ and enough blocks $m=O(D_T\log(2TD_T)+\log(T))$ to satisfy the constraint from \cref{prop:performance_reference_expert_no_restart}. Optimizing over the bank size gives the choice $N=\Theta(\sqrt{dT/m}+d)$. For the aggregation step, we have $\log L=O((dsM_T+1)\log(T+1))=O(d^2 D_T\log(2TD_T)\log^2(T+1))$. Altogether this gives the following result.

\begin{theorem}[Adaptive adversaries via direct aggregation]
\label{thm:adaptive-full-feedback-direct_no_phase}
There is a universal constant $C>0$ such that for every $T\geq 1$, $\epsilon\in(0,1]$, the following holds. Let
$\ell_T:=\lceil d\log(T+1)/\log(16/15)\rceil+1$ and suppose that $\mathcal F$ has
$\ell_T$-reduction dimension $D_T<\infty$.
Then \cref{alg:adaptive-direct-aggregation} run on the experts from \cref{sec:ac-scores} achieves for appropriate parameters:
\[
  \MisErr\le C\frac{d^2 D_T\log(2TD_T)\log^2(T+1)}{\epsilon},
  \qquad
  \E[\AbsErr]
  \le \epsilon T + C\sqrt{dD_T T\log(2D_T T)}.
\]
The mistake bound holds on every realized interaction.
\end{theorem}

Complete parameter calculation appears in \cref{subsec:final_proofs}.

\subsection{Completing the misclassification/abstention tradeoff}
\label{subsec:completing_tradeoff}

The main bottleneck in the previous algorithm came from the reference expert which abstains until it observes a large enough clean bank of size $Nm$. Instead of simply abstaining on this phase, we can use a similar reference expert which uses a smaller clean bank.
We repeat this idea over at most $J:=\ceil{\log_2 T}$ phases, using the compression to successively increase the clean bank size throughout the interaction. We refer to this as the \emph{multi-phase reference expert} as defined below.

\paragraph{Multi-phase reference expert.}
We can further remove this $Nm$ term using a recursive phase strategy---in each phase, the strategy uses clean rounds from previous phases as the clean bank.
The multi-phase process first abstains until it has observed $md$ clean instances. For $j\geq0$, let $n_j=md2^j$ and $N_j=d2^j$. Phase $j$ starts on round $\tau_j$ immediately after the $n_j$-th clean round. On each phase, it then uses the reference expert from \cref{sec:ac-scores} using the $n_j=mN_j$ clean bank already available, until the next phase starts. In particular, the version space is restarted at the beginning of each phase. We denote the number of phases by $h\leq J$. The corresponding multi-phase expert is summarized in \cref{alg:ac-candidate}.

\begin{algorithm}[tbp]
\caption{Multi-phase reference expert}
\label{alg:ac-candidate}
\begin{algorithmic}[1]
\Require Class $\mathcal F$, phase start-times $\tau_0<\cdots<\tau_{h-1} <\tau_h:=T+1$, witness-index lists $(\mathcal Z_{j,r})_{j<h,\,0\leq r\leq M_T}$
\State Abstain $\widehat y_t\gets \bot$ for all $t<\tau_0$
\State\textbf{For} each phase $j=0,\ldots,h-1$:
\State\quad \textbf{For} rounds $t\in\{\tau_j,\ldots,\tau_{j+1}-1\}$:
\State\quad\quad Run reference expert \cref{alg:reference_process_no_restart} for $\Fcal$, start time $\tau_j$, and witness-index list $(\mathcal Z_{j,r})_{0\leq r\leq M_T}$
\end{algorithmic}
\end{algorithm}

Note that on each phase $j$ the multi-phase reference expert coincides with the reference expert starting at time $\tau_j$. Hence, the misclassification and abstention error bounds from \cref{prop:performance_reference_expert_no_restart} still apply on each phase separately: the multi-phase expert makes at most $M_T$ mistakes on each phase and an abstention error on phase $j$ roughly $O(d(n_{j+1}-n_j)/N_j)=O(dm)$. Summing over all phases gives the following guarantee for the multi-phase reference expert:

\comment{
\paragraph{Clean abstentions during a stage.}
For a fixed nonempty version space, the one-block leave-one-out
bound is $2/T+d/(N+1)$. Reduction dimension and a constant-gap
VC bound make the one-eighth block test uniform over the version
spaces the expert may select. More precisely, if
\[
 g_W(S,x):=\ind\left\{
       \min_b\Theta^S(W^{x\to b})>\frac34\Theta^S(W)\right\},
\]
then $m$ independent clean blocks of size $N\leq T$ satisfy
\begin{equation}
 \mathbb E\left[
   \sup_{\substack{W\in\mathcal R_{\ell_T}\\W\ne\varnothing}}
    \mu\left\{x:\sum_{j=1}^m g_W(S_j,x)>m/8\right\}
          \right]
 \leq\frac{33}{T}+\frac{16d}{N+1}.
 \label{eq:ac-main-certificate}
\end{equation}
The current state may depend on the bank; the supremum is what
allows this dependence. Stage $j$ contains at most $n_j$ further
clean rounds, so its expected clean abstentions are at most
$n_j[33/T+16d/(N_j+1)]$. Summing over the doubling schedule
gives the following guarantee.
}

\begin{proposition}[Clean reference process]
\label{prop:ac-reference}
Let $\delta\in(0,1)$ and suppose that $\Fcal$ has $\ell_T$-reduction dimension $D_T$. If $m\geq C_m(D_T\log(2TD_T)+\log(2/\delta))$, then the multi-phase reference process has at most $JM_T$ mistakes on every
realized interaction and at most $\left(8dm + 32+\delta T\right)J +md$ 
expected abstention error.
\end{proposition}

Since the learner does not know a priori which are the phase start-times nor the valid witness-index lists, we consider the set of all experts for any choice of these parameters. Let $\Ccal'$ denote the resulting family. This family of experts has size $L$ with $\log L\leq(J+Jds(M_T+1))\log(T+1)=\tilde O(d^2 D_T)$. The learner then aggregates these learners using the same direct aggregation strategy from \cref{alg:adaptive-direct-aggregation}. The same choice of parameters as in \cref{sec:adaptive-candidate-aggregation} then gives the desired guarantee for \cref{thm:adaptive-full-feedback-direct}. Complete parameter calculation appears in \cref{subsec:final_proofs}.

\subsection{Extension to the censored feedback model}
\label{sec:adaptive-censored-extension}

As discussed in \cref{sec:adaptive-candidate-aggregation}, provided that the learner can simulate the candidate-bank experts on each round, the direct aggregation can be readily implemented in the censored model. However, the candidate-bank experts require knowing the label $y_t$ when they make a prediction mistake, which may not be observed by the aggregate learner. This can be resolved using the same arguments as in \cite{yu2026distributionfree}, by slightly modifying the candidate-bank experts as detailed below.

Up to increasing the number of experts, we may consider that the multi-phase reference experts are given their phase start-times $\{\tau_j\}_{j<h}$ and witness-index lists $(\mathcal Z_{j,r})_{j<h,\,0\leq r\leq M_T}$ together with a set of at most $J M_T$ rounds $S\subseteq [T]$ and corresponding labels $z_S\in\{0,1\}^S$. The expert then runs the previous multi-phase reference expert \cref{alg:ac-candidate} but only updates the version space $V$ at times $t\in S$ using the prescribed label $z_t$. Formally, if $t\in S$ it updates $V\gets V^{x_t\to z_t}$. In particular, the new experts do not require observed labels at all and hence can be implemented in the censored model. For convenience, denote by $\widetilde{\mathcal C}$ the collection of all these experts. Due to the choice of rounds $S$ and labels $z_S$ we have
\begin{equation*}
    |\widetilde {\mathcal C}|\leq |\mathcal C'| (2T+1)^{J M_T}.
\end{equation*}

Since the matching multi-phase reference expert from \cref{prop:ac-reference} makes at most $J M_T$ mistakes, one of the previous censored experts emulates it, enjoying the same mistake and abstention guarantee. We can then apply directly the guarantee from the direct aggregation algorithm \cref{prop:adaptive-direct-aggregation-guarantee} with the only difference that $L=|\widetilde {\mathcal C}|$. This increases the term $R$ by an extra $J M_T \log(2T)$ additive term as compared to the full-feedback case. With the parameters above, $J M_T\log(2T)=O(\log|\mathcal C'|)$. Hence, this modification only affects the results up to a constant factor, proving that \cref{thm:adaptive-full-feedback-direct} holds in the censored feedback as well.

\section{Conclusion}
\label{sec:conclusion_new}

Under the sequential learning with abstention framework, we propose several weighted-shattering-potential-based algorithms and improve the results from \cite{goel2023adversarial} and \cite{yu2026distributionfreev2}. When the clean distribution is known, our algorithm achieves the optimal $O(d\log T)$ rate. When the distribution is unknown, we improve the abstention-mistake tradeoff in \cite{yu2026distributionfreev2} for both oblivious and adaptive adversary, under both censored and full feedback. In particular, we close gaps up to constant factors for oblivious adversaries.
The following questions remain open:


\paragraph{Extending results to adaptive adversaries for general VC classes.} As in \cite{yu2026distributionfreev2}, our guarantees for adaptive adversaries are limited to finite-VC classes with finite reduction dimension. It remains open whether one can learn all finite-VC classes without additional structural assumptions. 


\paragraph{Computationally efficient abstention learning.} All algorithms we propose in this paper are information-theoretic. In particular, computing the weighted shattering potential would typically require a runtime exponential in the dimension $d$. A natural question would be whether there exist efficient algorithms for this abstention learning problem, for instance using the ERM-oracle efficiency model 
\cite{hazan2016computational,dudik2020oracle,assos2023online,kalai2021optimally,haghtalab2022oracle,wu2024oracle}.


\paragraph{Beyond i.i.d.\ clean samples.} Our analysis for oblivious adversaries and uncensored feedback works under the weaker assumption that clean round samples are \emph{exchangeable}. A natural question is whether the analysis for the censored case can also be carried out under exchangeability alone. Similarly, for adaptive adversaries one could consider a stronger adversarial model in which an adversary first selects the clean sample pool but not their random appearance order, then adaptively selects whether to corrupt or not each round. 

\subsection*{Acknowledgments}
SG gratefully acknowledges the support from NSF Award CCF:AF:2504016. 

\subsection*{Statement on the use of AI}
 The authors used GPT 5.6 Pro and GPT 6 Astra during the later stages of the project to improve parts of the adaptive full-feedback and oblivious censored-feedback proofs. The main results and initial proofs were developed by the authors prior to this use. The authors have reviewed and verified all AI-assisted proofs and writing and take full responsibility for the final contents of the paper.
\bibliographystyle{plainnat}
\bibliography{refs}

\appendix
\section{Unknown-Distribution Proofs}
\label{app:unknown-proofs}

We prove the upper and lower bounds for an unknown clean distribution under oblivious adversaries and full feedback. This section is organized as follows. In \cref{app:unknown-upper-proofs} we prove the misclassification and abstention guarantee for the algorithm using the simpler weighted history potential (see \cref{def:empirical-potential}). We give a deterministic mistake guarantee, then a replace-one argument
bounds the number of bad positions in any realized labeled sequence, where a position is bad if
the same rule would abstain when that position is replayed last.  Exchangeability turns this count
into a clean-round abstention probability. Next, in \cref{app:unknown-upper-proofs-age-weighted} we adapt the proof structure to the age-weighted history potential. The mistake guarantee extends immediately, while the exchangeability argument requires additional care to bound certain fractions of partition functions.
Finally, in \cref{app:unknown-lower-proof} we prove the lower bound. To do so, we lift the one-dimensional
construction of \citet{edelman2026reliable} to VC dimension $d$ by a direct sum and derive the resulting mistake--abstention tradeoff.

\subsection{Upper bound for the Weighted History Potential}
\label{app:unknown-upper-proofs}

\begin{proof}[Proof of \cref{lem:empirical-growth}]
Let $S$ have length $n$, so the appended point has index $n+1$.  Every
$J\in\Sh(S\Vert z)$ falls into one of two cases.  If $n+1\notin J$, then $J\in\Sh(S)$: deleting
the new consistency constraint can only make shattering easier.  If $n+1\in J$, set
$I:=J\setminus\{n+1\}$.  Then $I\in\Sh(S)$, since a labeling of $I$ can be realized by invoking
the shattering of $J$ with the new coordinate fixed, say, to zero.  Therefore
\begin{align*}
  \widehat\Theta_\beta(S\Vert z)
  &=\sum_{J\in\Sh(S\Vert z)}\beta^{|J|}\\
  &\leq
    \sum_{J\in\Sh(S)}\beta^{|J|}
    +\sum_{I\in\Sh(S)}\beta^{|I|+1}
   =(1+\beta)\widehat\Theta_\beta(S).
\end{align*}
\end{proof}

\begin{proof}[Proof of \cref{prop:empirical-mistakes}]
Set $\Phi_t:=\widehat\Theta_\beta(S_t)$ for $t=0,\ldots,T$.  Every prefix is realizable by
$f^\star$, so $\varnothing\in\Sh(S_t)$ and $\Phi_t\geq1$.  Moreover,
$\Phi_0=\widehat\Theta_\beta(())=1$.  By \cref{lem:empirical-growth}, every round satisfies
\[
  \Phi_t\leq(1+\beta)\Phi_{t-1}.
\]
On a mistake round, the learner predicted a branch of maximum potential and the update follows the
other branch, which has minimum potential.  Since the learner predicted rather than abstained,
\[
  \Phi_t=\min\{P_0,P_1\}\leq\tfrac34\Phi_{t-1}.
\]
Writing $M=\MisErr$ and iterating gives
\[
  1\leq\Phi_T
  \leq\left(\frac34\right)^M(1+\beta)^{T-M}.
\]
Taking logarithms yields
\[
  M
  \leq
  \frac{T\log(1+\beta)}{\log(4(1+\beta)/3)}.
\]
\end{proof}

\begin{proof}[Proof of \cref{lem:empirical-pointwise}]
Fix $i\in[n]$.  For each $I\subseteq[n]\setminus\{i\}$ and $y\in\{0,1\}$, define
\[
  a_y(I):=\1\{I\in\Sh(S[i\leftarrow y])\},
  \qquad
  b(I):=\1\{I\in\Sh(S_{-i})\},
  \qquad
  q(I):=\1\{I\cup\{i\}\in\Sh(S)\}.
\]
For every such $I$,
\begin{equation}
  a_0(I)+a_1(I)\leq b(I)+q(I).
  \label{eq:pointwise-indicator-decomposition}
\end{equation}
Indeed, if $a_y(I)=1$ for either $y$, deleting the constraint at $i$ can only help, so $b(I)=1$.
If both $a_0(I)=a_1(I)=1$, then any labeling of $I\cup\{i\}$ can be realized by choosing a witness
from $S[i\leftarrow0]$ or $S[i\leftarrow1]$ according to the desired label at $i$; hence
$q(I)=1$.

Decompose $P_i(y)$ according to whether the shattered set contains $i$.  If a set contains $i$, its
membership in $\Sh(S[i\leftarrow y])$ is independent of the stored label at $i$, and hence is the same as
its membership in $\Sh(S)$.  Therefore
\[
  P_i(y)
  =\sum_{I\subseteq[n]\setminus\{i\}}\beta^{|I|}a_y(I)
   +\beta W_i(S).
\]
Summing over $y$ and using \eqref{eq:pointwise-indicator-decomposition} yields
\begin{align*}
  P_i(0)+P_i(1)
  &\leq
    \sum_{I\subseteq[n]\setminus\{i\}}
       \beta^{|I|}\bigl(b(I)+q(I)\bigr)
    +2\beta W_i(S)\\
  &=\widehat\Theta_\beta(S_{-i})+W_i(S)+2\beta W_i(S),
\end{align*}
where
$\sum_I\beta^{|I|}q(I)=W_i(S)$.  This is the claimed inequality.
\end{proof}

\begin{proof}[Proof of \cref{prop:few-bad-indices}]
Fix $i\in\operatorname{Bad}(S)$.  Badness and \cref{lem:empirical-pointwise} imply
\[
  \frac32\widehat\Theta_\beta(S_{-i})
  <P_i(0)+P_i(1)
  \leq\widehat\Theta_\beta(S_{-i})+(1+2\beta)W_i(S),
\]
and consequently
\begin{equation}
  W_i(S)>
  \frac{\widehat\Theta_\beta(S_{-i})}{2(1+2\beta)}.
  \label{eq:bad-index-local-weight}
\end{equation}
The sequence $S$ is a permutation of $S_{-i}\Vert z_i$.  Permutation invariance and
\cref{lem:empirical-growth} therefore give
\[
  \widehat\Theta_\beta(S)
  \leq(1+\beta)\widehat\Theta_\beta(S_{-i}).
\]
Combining with \eqref{eq:bad-index-local-weight}, every bad index satisfies
\[
  W_i(S)>
  \frac{\widehat\Theta_\beta(S)}{2(1+2\beta)(1+\beta)}.
\]
Summing this strict inequality over the bad indices gives
\begin{equation}
  \frac{|\operatorname{Bad}(S)|}{2(1+2\beta)(1+\beta)}
  \widehat\Theta_\beta(S)
  <\sum_{i\in\operatorname{Bad}(S)}W_i(S)
  \leq\sum_{i=1}^nW_i(S).
  \label{eq:bad-index-sum-lower}
\end{equation}
On the other hand, every shattered set has size at most $d$, so
\begin{equation*}
\begin{aligned}
  \sum_{i=1}^nW_i(S)
  &=\sum_{U\in\Sh(S)}|U|\beta^{|U|-1}\\
  &\leq\frac d\beta\sum_{U\in\Sh(S)}\beta^{|U|}
   =\frac d\beta\widehat\Theta_\beta(S).
\end{aligned}
\end{equation*}
Since $S$ is realizable, $\widehat\Theta_\beta(S)\geq1$.  Substituting the preceding upper bound on
$\sum_{i=1}^n W_i(S)$ into \eqref{eq:bad-index-sum-lower} and canceling the positive potential gives
\[
  |\operatorname{Bad}(S)|
  \leq\frac{2d(1+2\beta)(1+\beta)}{\beta}.
\]
\end{proof}

\begin{proof}[Proof of \cref{prop:clean-round-exchangeability}]
Fix the adversary's private seed.  The clean rounds
$1\leq s_1<\cdots<s_N\leq T$ and all corrupted examples are now fixed.  Let
\[
  Z_j:=(x_{s_j},f^\star(x_{s_j})),
  \qquad j\in[N].
\]
Because the adversary is oblivious, $Z_1,\ldots,Z_N$ are i.i.d.\ labeled draws.  For $i\in[N]$, let
$C_i$ be the labeled sequence of corrupted rounds preceding $s_i$ and form the reordered sequence
\[
  R_i:=C_i\Vert Z_1\Vert\cdots\Vert Z_i.
\]
The actual history immediately before round $s_i$ is a permutation of $R_i$ with its final entry
removed.  Since the weighted history potential is permutation invariant, the learner abstains at $s_i$
exactly when the final index of $R_i$ is bad:
\begin{equation}
  \Pp(\widehat y_{s_i}=\bot)
  =\Pp\bigl(|R_i|\in\operatorname{Bad}(R_i)\bigr).
  \label{eq:last-clean-is-bad}
\end{equation}

The prefix $C_i$ is fixed, while the i.i.d.\ clean examples fill the last $i$ positions.  Permuting
these positions leaves the law of $R_i$ unchanged.  It also permutes the membership of the clean
positions in $\operatorname{Bad}(R_i)$, because deletion, relabeling, and
$\widehat\Theta_\beta$ are equivariant under reordering.  All clean positions are therefore equally
likely to be bad, and
\begin{align*}
  \Pp\bigl(|R_i|\in\operatorname{Bad}(R_i)\bigr)
  &=\frac1i\E\bigl[
      |\operatorname{Bad}(R_i)\cap\{\text{clean positions}\}|
    \bigr]\\
  &\leq\frac1i\E[|\operatorname{Bad}(R_i)|]\\
  &\leq\frac{2d(1+2\beta)(1+\beta)}{\beta i},
\end{align*}
where the last step is \cref{prop:few-bad-indices}.  Combine with
\eqref{eq:last-clean-is-bad}.
\end{proof}

We are now ready to prove a misclassification and abstention error bound for the algorithm using the weighted history potential $\widehat\Theta_\beta$.

\begin{theorem}[Unknown distribution, oblivious corruption]
\label{thm:unknown-oblivious}
For every $\beta>0$, \cref{alg:empirical} with non-age-dependent potential $\widehat\Theta_\beta$ satisfies, on every realizable labeled interaction,
\begin{equation}
   \MisErr
   \leq
   \frac{\log(1+\beta)}{\log(4(1+\beta)/3)}\,T.
 \end{equation}
 Against an oblivious adversary,
 \begin{equation}
   \E[\AbsErr]
   \leq
   \frac{2d(1+2\beta)(1+\beta)}{\beta}\,H_T.
 \end{equation}
 The expectation is over the clean draws and any randomization of the adversary and learner.
 \end{theorem}
\begin{proof}
The mistake claim is \cref{prop:empirical-mistakes}.  For abstentions, condition on the oblivious
adversary's private seed and retain the notation from the preceding proof.  If $N=0$, then
$\AbsErr=0$ and there is nothing to prove.  Otherwise,
\cref{prop:clean-round-exchangeability} gives
\begin{align*}
  \E[\AbsErr\mid\text{seed}]
  &\leq
  \frac{2d(1+2\beta)(1+\beta)}{\beta}
  \sum_{i=1}^{N}\frac1i\\
  &=\frac{2d(1+2\beta)(1+\beta)}{\beta}H_N\\
  &\leq\frac{2d(1+2\beta)(1+\beta)}{\beta}H_T.
\end{align*}Taking expectation over the adversary's seed proves the claimed abstention bound.
\end{proof}

\subsection{Upper bound for the Age-Weighted History Potential}
\label{app:unknown-upper-proofs-age-weighted}

Bounding the misclassification error for this potential follows the same template as for the simpler potential $\widehat \Theta_\beta$. The following is the direct analog of \cref{lem:empirical-growth}.

\begin{lemma}[One-step growth]
\label{lemma:bounding-potential-growth-remove-log}
    For every labeled history $S$ and any new labeled example $z$,
    \begin{equation*}
        \widehat\Theta_{\lambda}(S\parallel z)\leq e^{2\beta}\widehat\Theta_{\lambda}(S).
    \end{equation*}
\end{lemma}

\begin{proof}
    Since $z$ is added last, its age-dependent weight within the new dataset $S\parallel z$ is $\beta$. Additionally, note that for any $u\leq t$ one has
    \begin{equation*}
        \frac{\lambda_{u,t+1}}{\lambda_{u,t}} \leq 1+\frac{\beta}{d}.
    \end{equation*}
    We then compute
    \begin{align*}
        \widehat\Theta_{\lambda}(S\parallel z) &\leq \sum_{U\in\Sh(S)} \prod_{u\in U}\lambda_{u,t+1} + \sum_{U\subseteq[n]: U\cup\{z\}\in \Sh(S\parallel z)} \beta \prod_{u\in U}\lambda_{u,t+1}\\
        &\leq (1+\beta) \sum_{U\in\Sh(S)} \prod_{u\in U}\lambda_{u,t+1}\\
        &\leq (1+\beta)\sum_{U\in\Sh(S)} \paren{1+\frac{\beta}{d}}^{|U|}\prod_{u\in U}\lambda_{u,t} \leq (1+\beta)\paren{1+\frac{\beta}{d}}^{d} \widehat\Theta_{\lambda}(S).
    \end{align*}
    Using $1+x\leq e^x$ ends the proof.
\end{proof}

From there, the same arguments give the following misclassification bound. Specifically, if $M$ mistakes have been made in total and $S_T$ denotes the dataset after all $T$ rounds, we have
\begin{equation*}
    1\leq \widehat\Theta_\lambda(S_T) \leq e^{2\beta(T-M)}\paren{\frac{3}{4}}^M.
\end{equation*}
This corresponds to the following misclassification bound:

\begin{proposition}
\label{prop:empirical-mistakes-age-potential}
For every realizable interaction, \cref{alg:empirical} with the age-dependent potential satisfies
\[
  \MisErr
  \leq
  \frac{2\beta}{\log(4/3)+2\beta}\,T.
\]
\end{proposition}

We next turn to bounding the abstention error with the new potential. The end goal of this part is to prove the following abstention probability on the $i$-th clean sample:

\begin{proposition}
\label{lemma:clean-round-exchangeability-log-removed}
For the $i$th clean round $s_i$ of any fixed oblivious corruption schedule,
\[
  \Pp(\widehat y_{s_i}=\bot)
  \lesssim \paren{\frac{d}{\beta i}}^2 + \frac{d}{i}.
\]
\end{proposition}We defer the proof to \cref{subsec:abstention_probability_bound}. Compared to \cref{prop:clean-round-exchangeability} which gave a bound $O(\frac{d}{\beta i})$ for the simpler potential $\widehat\Theta_\beta$, \cref{lemma:clean-round-exchangeability-log-removed} shows a tighter tail decay for the age-weighted potential $\widehat\Theta_\lambda$ ($\sim 1/i^2$ instead of $\sim 1/i$). This in turn removes a $\log T$ factor for the abstention error once these probabilities are summed.

\begin{proposition}
\label{prop:empirical-abstention-age-potential}
Fix $\beta\in(0,1/4]$. Against any oblivious adversary, \cref{alg:empirical} with the age-dependent potential satisfies
\[
  \mathbb E[\AbsErr]
  \lesssim
  \frac{d}{\beta} + d\log \paren{1+\frac{\beta T}{d}}.
\]
\end{proposition}

\begin{proof}
    The expected abstention error is exactly the sum of probabilities of abstention for each clean round, which are bounded in \cref{lemma:clean-round-exchangeability-log-removed}. Hence, letting $N$ be the number of clean rounds,
    \begin{equation*}
        \mathbb E[\AbsErr] \lesssim \sum_{i=1}^N \min\paren{\paren{\frac{d}{\beta i}}^2 + \frac{d}{i},1} \lesssim \frac{d}{\beta} + d\log \paren{1+\frac{\beta N}{d}}.
    \end{equation*}
    Using $N\leq T$ ends the proof.
\end{proof}

Together, \cref{prop:empirical-mistakes-age-potential,prop:empirical-abstention-age-potential} prove \cref{thm:unknown-oblivious-no-log}.

\subsubsection{Bounding abstention probabilities}
\label{subsec:abstention_probability_bound}

For convenience, we introduce a few definitions. Given a dataset $S\subseteq \X\times\{0,1\}$, and weights $\lambda:S\to\R_+$, for any family $T\subseteq 2^S$ we define the corresponding partition function
\begin{equation*}
    Z_{\lambda}(T):=\sum_{U\in T} \prod_{u\in U}\lambda(u).
\end{equation*}
In particular, if we denote by $\lambda_t$ the age-dependent weights, then the age-dependent potential is obtained by taking the partition function over all shattered sets in $S$: $\widehat\Theta_\lambda(S)=Z_{\lambda_t}(\Sh(S))$. 

We will consider two different families. \emph{Old} shattered sets correspond to those that are shattered without using the last datapoint: 
\begin{equation*}
    \Ocal:=\{U\in\Sh(S): t\notin U\} \subseteq \Sh(S_{-t}).
\end{equation*}
In particular, $\Ocal\subseteq\Sh(S_{-t})$.
Conversely, \emph{new} shattered sets are those that were created when adding the new datapoint:
\begin{equation*}
    \Ncal:=\{U\subseteq[t-1]: U\cup\{t\}\in\Sh(S)\}.
\end{equation*}

We start by observing that the decomposition from the previous potential $\widehat\Theta_\beta$ still applies here up to minor modifications. By slight abuse of notation, we denote by $P_t(0)$ and $P_t(1)$ the two next-round potentials if the label is $0$ or $1$. In particular, the learner abstains if
\begin{equation*}
    \min_{y\in\{0,1\}} P_t(y) > \frac{3}{4}\widehat\Theta_{\lambda,t-1}(S_{-t}).
\end{equation*}

Similar arguments as in \cref{lem:empirical-pointwise} up to book-keeping weights show the following identity when a new point triggers an abstention.

\begin{lemma}\label{lemma:rewriting-abstention-event}
    Let $\beta\in(0,1/4]$. For every history $S=(x_u,y_u)_{u<t}$ and new point $(x_t,y_t)$, if $x_t$ triggered an abstention, then denoting by $\lambda_t$ the age-dependent weights,
    \begin{equation}
        Z_{\lambda_t}(\Ncal)  \geq \frac{1}{10} Z_{\lambda_t}(\Ocal).
    \end{equation}
\end{lemma}

\begin{proof}
    The same arguments as in \cref{lem:empirical-pointwise} up to book-keeping weights give the following similar potential decomposition:
    \begin{equation}\label{eq:empirical-pointwise-log-factors}
        P_t(0)+P_t(1)
  \leq
  Z_{\lambda_t}(\Sh(S_{-t}))+(1+2\beta)Z_{\lambda_t}(\Ncal).
    \end{equation}

    Next, recall that for any $u<t$ one has $\lambda_{u,t}\leq (1+\beta/d)\lambda_{u,t-1}$. As a result, when a learner abstains one has
    \begin{equation*}
        P_t(0)+P_t(1) \geq \frac{3}{2}\widehat\Theta_{\lambda_{t-1}}(S_{-t}) \geq \frac{3}{2(1+\beta/d)^d} Z_{\lambda_t}(\Sh(S_{-t})) \geq \frac{3}{2e^\beta} Z_{\lambda_t}(\Sh(S_{-t}))
    \end{equation*}
    Together with \cref{eq:empirical-pointwise-log-factors} this implies
    \begin{equation*}
        Z_{\lambda_t}(\Ncal) \geq \frac{\frac{3}{2}e^{-\beta}-1}{1+2\beta}Z_{\lambda_t}(\Sh(S_{-t})) \geq \frac{1}{10}Z_{\lambda_t}(\Sh(S_{-t})).
    \end{equation*}
    In the last inequality we used $\beta\leq 1/4$. Noting that $\Ocal\subseteq \Sh(S_{-t})$ ends the proof.
\end{proof}

Finally, we use an exchangeability argument to bound the probability of this event happening. This is the main step of the proof.
Denote the number of clean rounds by $N$. In the following, we fix $i\in[N]$ and let $\tau_i$ denote the time of the $i$-th clean round. The following result bounds the probability of the event $\{Z_{\lambda_t}(\Ncal) \geq\frac{1}{10}Z_{\lambda_t}(\Ocal)\}$ when randomly rotating the order of clean samples arrival.

Fix corrupted samples on all rounds and also fix the instances within clean rounds: $\{x_{\tau_j}:j\in[i]\}$ but not their assignment to clean rounds $C_i:=\{\tau_j:j\leq i\}$. We have the following abstention probability bound where the probability is taken only over this random assignment. This immediately implies \cref{lemma:clean-round-exchangeability-log-removed}.

\begin{proposition}\label{prop:exchangeability_argument}
    Let $i\in[N]$. Fix all corrupted instances, as well as clean round instances before $\tau_i$ (but not their assignment to clean rounds). Let $\pi\in \mathrm{Sym}(C_i)$ be a uniform reordering of the clean rounds. We denote by $S_i^\pi=(x_{\pi(u)},y_{\pi(u)})_{u\leq t}$ the corresponding reordered dataset. Similarly, we denote by $\Ncal_i^\pi$ and $\Ocal_i^\pi$ the corresponding old and new shattered set families, and $\lambda^\pi$ the corresponding age-dependent weights. Then, for $\beta\in(0,1/4]$ and any universal constant $c\in (0,1]$,
    \begin{equation*}
        \mathbb P_\pi\sqb{Z_{\lambda^\pi}(\Ncal_i^\pi) \geq c Z_{\lambda^\pi}(\Ocal_i^\pi)} \lesssim \frac{d}{i}+ \paren{\frac{d}{\beta i}}^2.
    \end{equation*}
\end{proposition}

\begin{proof}
    For convenience, we use the notation $t=\tau_i$, and denote by $S_0:=\{(x_{\tau_j},y_{\tau_j}):j\in[i]\}$ the clean instances that have been fixed (without their assignment to clean slots), and $S=\{(x_u,y_u), u\leq t\}$ the complete dataset. For any clean datapoint $v\in S_0$, we introduce
    \begin{align*}
        \Ocal(v)&:= \{ U\in \Sh(S): v\notin U\}\\
        \Ncal(v)&:= \{ U\subseteq S\setminus\{v\}: U\cup\{v\}\in \Sh(S)\}.
    \end{align*}
    In particular, for a fixed assignment $\pi$, denoting by $e=e(\pi)$ the corresponding last datapoint observed at time $t$, we have $\Ocal_i^\pi = \Ocal(e)$ and $\Ncal_i^\pi=\Ncal(e)$. Last, we introduce distributions $\mu_\pi$ and $\nu_\pi$ corresponding to the distribution on $\Sh(S)$ and $\Ncal(e)$ corresponding to the partition function:
    \begin{equation*}
        \mu_\pi(U):=\frac{\prod_{u\in U} \lambda^\pi(u)}{Z_{\lambda^\pi}(\Sh(S))},\quad U\in\Sh(S).\quad\text{and} \quad \nu_\pi(U):=\frac{\prod_{u\in U} \lambda^\pi(u)}{Z_{\lambda^\pi}(\Ncal(e))},\quad U\in\Ncal(e).
    \end{equation*}
    For any clean datapoint $v\in S_0$, we denote by $p_v(\pi)$ and $q_v(\pi)$ the probability that $v$ belongs to a random shattered set according to $\mu_\pi$ and $\nu_\pi$ respectively:
    \begin{equation*}
        p_v(\pi):=\mathbb P_{U\sim\mu_\pi}[v_r\in U] \quad\text{and} \quad q_v(\pi):=\mathbb P_{U\sim\nu_\pi}[v_r\in U].
    \end{equation*}
    In particular, since shattered sets contain at most $d$ points, we have
    \begin{equation}\label{eq:max_nb_elements_shattered}
        \sum_{v\in S_0}p_v(\pi) \leq\mathbb E_{U\sim\mu_\pi}[|U|]\leq d \quad\text{and} \quad \sum_{v\in S_0} q_v(\pi)\leq d.
    \end{equation}
    We suppose in the following that the assignment $\pi$ satisfies the considered property
    \begin{equation}\label{eq:partition_comparison}
        Z_{\lambda^\pi}(\Ncal_i^\pi) \geq c Z_{\lambda^\pi}(\Ocal_i^\pi).
    \end{equation}

    Fix $r\in[i]$ and denote by $v_r$ the datapoint assigned to the $r$-th clean round in $\pi$. We consider the assignment $\pi^{(r)}$ obtained by swapping the clean observation in the $r$-th clean round with the last $i$-th clean round (corresponding to time $t=\tau_i$). Under $\pi^{(r)}$, the age-dependent weight of $v_r$ is reduced to $\beta$ while the weight of $e$ becomes
    \begin{equation*}
        a_r:=\beta\paren{1+(t-\tau_r)\frac{\beta}{d}} \geq\beta.
    \end{equation*}
    All other age-dependent weights are kept unchanged. In particular, all sets in $\Ncal(e)$ which didn't include $v_r$ have the same weight under $\pi$ and $\pi^{(r)}$. Therefore,
    \begin{align*}
        Z_{\lambda^{\pi^{(r)}}}(\Ncal(e)) \geq \mathbb P_{U\sim \nu_\pi}[v_r\notin U] \cdot Z_{\lambda^\pi}(\Ncal(e))
        &= (1-q_{v_r}(\pi))Z_{\lambda^\pi}(\Ncal(e))\\
        &\overset{(i)}{\geq} c(1-q_{v_r}(\pi)) Z_{\lambda^\pi}(\Ocal(e))\\
        &\overset{(ii)}{\geq} c(1-q_{v_r}(\pi)) Z_{\lambda^{\pi^{(r)}}}(\Ocal(e)).
    \end{align*}
    In $(i)$ we used the assumption \cref{eq:partition_comparison} and in $(ii)$ we used the fact that old shattered sets can only have reduced weight when the datapoint $v_r$ is brought closer to the end time $t$.

    We now turn our attention to the probability $p_e(\pi^{(r)})$ that $e$ belongs to the distribution on $Sh(S)$ induced by $\pi^{(r)}$. Since $e$ has weight $a_r$ under $\pi^{(r)}$ we have
    \begin{align*}
        p_e(\pi^{(r)})&=\frac{a_r  Z_{\lambda^{\pi^{(r)}}}(\Ncal(e))}{Z_{\lambda^{\pi^{(r)}}}(\Ocal(e)) +  a_r Z_{\lambda^{\pi^{(r)}}}(\Ncal(e))}\\
        &\overset{(i)}{\geq} \frac{ca_r(1-q_{v_r}(\pi))}{1+ca_r(1-q_{v_r}(\pi))}\\
        &\overset{(ii)}{\geq} \frac{1}{2}\min\paren{ca_r (1-q_{v_r}(\pi)),1},
    \end{align*}
    where in $(i)$ we used the previous inequality and in $(ii)$ we used $\frac{x}{1+x}\geq \frac{1}{2}\min(x,1)$ for $x\geq 0$. Summing these inequalities over $r$ then gives
    \begin{align*}
        2\sum_{r\in[i]} p_e(\pi^{(r)}) &\geq \sum_{r\in[i]} \min(ca_r,1) (1-q_{v_r}(\pi))\\
        &\geq c\sum_{r\in[i]} \min(a_r,1) - \sum_{r\in[i]} q_{v_r}(\pi)\\
        &\overset{(i)}{\geq} c\sum_{r\in[i]} \min(a_r,1) - d.
    \end{align*}
    In $(i)$ we used \cref{eq:max_nb_elements_shattered}. As an important reminder, this bound holds under the event from \cref{eq:partition_comparison}. On the other hand,
    \begin{equation*}
        \sum_{r\in[i]} \min(a_r,1) \geq \sum_{r\in[i]} \min\paren{\beta\paren{1+(i-r)\frac{\beta}{d}},1} \gtrsim i\cdot \min\paren{\beta + \frac{\beta^2(i-1)}{d}, 1}.
    \end{equation*}

    Importantly, the mapping $(\pi,r)\to(\pi^{(r)},e)$ is injective: given a target $(\sigma,v)$, $\pi$ is recovered by swapping $v$ with the $t$-th datapoint of $\sigma$, and $r$ is the index of $v$ within $\sigma$. As a result, the last inequality implies that if $\sigma$ is a uniform assignment,
    \begin{align*}
        \mathbb E_\sigma \sqb{\sum_{r\in[i]} p_e(\sigma^{(r)})}\leq  \mathbb E_\sigma \sqb{\sum_{v\in S_0} p_v(\pi)} \leq d,
    \end{align*}
    where in the last inequality we used \cref{eq:max_nb_elements_shattered}. Taking the last three inequalities together implies that for some universal constant $c'>0$, 
    \begin{equation*}
        \mathbb P_\sigma[Z_{\lambda^\sigma}(\Ncal_i^\sigma) \geq c Z_{\lambda^\sigma}(\Ocal_i^\sigma)] \leq \frac{2d}{c' i\cdot \min\paren{\beta + \frac{\beta^2(i-1)}{d}, 1} - d}.
    \end{equation*}
    For $i\lesssim d/\beta$ we can simply bound the probability by one. For $i\gtrsim d/\beta$ the bound above yields
    \begin{equation*}
        \mathbb P_\sigma[Z_{\lambda^\sigma}(\Ncal_i^\sigma) \geq c Z_{\lambda^\sigma}(\Ocal_i^\sigma)] \lesssim \frac{d}{i\min\paren{\frac{\beta^2 i}{d}, 1} }\lesssim \frac{d}{i} + \frac{d^2}{\beta^2 i^2}.
    \end{equation*}
    This ends the proof.
\end{proof}

\subsection{The direct-sum lower bound}
\label{app:unknown-lower-proof}

\begin{proof}[Proof of \cref{thm:unknown-lower-bound}]
We prove the budget tradeoff; its balanced specialization gives the
$\Omega(\sqrt{dT})$ total-cost bound.  The hidden paths, the identity of the clean block,
and all corrupted examples are sampled at time zero.  Once these objects are fixed, the
construction is an ordinary oblivious clean-label instance with one fixed clean distribution; only
the fresh clean draws remain random.
Throughout the proof, expectations include the hard-instance distribution, the fresh clean draws,
and any internal randomness of the learner.  The abstention identity and mistake inequality below
hold directly with all three sources of randomness included.  At the end, a dichotomy at the level
of the hard family selects one fixed member witnessing one of the theorem's two alternatives.

\paragraph{Step 1: the class.}
Set
\[
  m:=\min\{d,A\}.
\]
Because $d$ and $A$ are positive integers, $1\leq m\leq d$.  Fix integers $b\geq2$ and
$L\geq1$.  We use the rooted $b$-ary tree-path class appearing in the VC-$1$ lower bounds of
\citet{edelman2026reliable} and \citet{yu2026distributionfreev2}, truncated at depth $L$: a
hypothesis chooses a root-to-leaf path and labels exactly its nodes by one.  The
hidden-clean-block sampling below is the Edelman--Goel construction; Yu--Blanchard use the same
type of VC-$1$ tree class with a different recursive adversary.  Either source establishes that
one copy has VC dimension one.  For completeness, no two nodes are shattered: incomparable nodes
cannot both be positive, while for comparable nodes the labeling that makes the ancestor negative
and the descendant positive is impossible; any non-root node is itself shattered.

Take $m$ disjoint active copies and let a hypothesis choose its path in each copy independently.
Their direct sum has VC dimension exactly $m$.  To obtain dimension exactly $d$, take the
product with the class of all labelings of $d-m$ additional isolated points.  These additional
points never appear in the process, so this padding changes neither the transcript nor either loss.
The resulting class has VC dimension $m+(d-m)=d$.

For the tradeoff below, the displayed finite class may depend on $T$ and $A$ through $b$ and
$L$.  Equivalently, one can fix the active tree class once and for all using $m$ disjoint
countably infinite rooted trees and realize every finite construction inside them.

\paragraph{Step 2: the schedule and the hidden clean block.}
In active copy $r\in[m]$, let $v_{r,j}$ be the hidden path node at depth $j$, with
$v_{r,0}$ the root.  At each depth $j$, choose $v_{r,j}$ independently and uniformly among
the $b$ children of $v_{r,j-1}$.  For $j\in[L]$, let $P_j$ be uniform over all children of
the current depth-$(j-1)$ path nodes across the $m$ active copies:
\[
  P_j:=\Unif\bigl\{
      u:u\text{ is a child of }v_{r,j-1}\text{ for some }r\in[m]
    \bigr\}.
\]
Choose a hidden clean block $J\sim\Unif([L])$ and set the clean distribution to
$\mu:=P_J$.  Split the first
\[
  T_0:=Lmb
\]
rounds into $L$ blocks of length $mb$.  In block $j$, if $j=J$, all rounds are clean
i.i.d.\ draws from $P_j$; otherwise every round is corrupted and its instance is drawn from
$P_j$ at time zero.  All labels are the truthful labels induced by the same hidden paths.

Whenever $T_0<T$, pad the remaining $T-T_0$ rounds with corrupted repetitions of the root of
the first active tree, which is labeled one by every target in the active class.  This padding is
oblivious and clean-label.  Abstentions on the padded rounds are free, and any additional mistakes
only increase the lower-bound cost, so it is enough to analyze the first $T_0$ rounds.  This also
specifies a full horizon-$T$ instance rather than silently discarding unused rounds.

Under the distributional construction, the labeled transcript has the same law for every value of
$J$: block $j$ always consists of $mb$ draws from $P_j$ with the labels of the same hidden
paths.  The learner never observes the corruption indicators.  Let $N_\bot(j)$ be its number of
abstentions in block $j$ and let
\[
  N_\bot:=\sum_{j=1}^L N_\bot(j)
\]
be its total number of abstentions in the first $T_0$ rounds.  Since only block $J$ is clean and
$J$ is uniform,
\begin{align}
  \E[\AbsErr]
  &=\frac1L\sum_{j=1}^L\E[N_\bot(j)\mid J=j]
    =\frac1L\sum_{j=1}^L\E[N_\bot(j)]
    =\frac1L\E[N_\bot].
  \label{eq:lower-abstention-identity}
\end{align}

\paragraph{Step 3: uncertain rounds.}
In block $j$, each draw first selects an active copy $r$ and then a child of the current path
node $v_{r,j-1}$.  Call the round \emph{uncertain} if it is the first occurrence of the selected
child, the positive child of that parent has not appeared earlier in the block, and at least two
children remain possible before this occurrence.  We now extend the one-copy coupon calculation
of \citet{edelman2026reliable} to the uniform mixture of the $m$ active copies.

Fix one block and first expose the copy index of each of its $mb$ draws.  For active copy
$r\in[m]$, let $N_r$ be its number of draws, and list their child indices in within-copy order as
$C_{r,1},\ldots,C_{r,N_r}\in[b]$.  Then
\[
  N_r\sim\operatorname{Bin}(mb,1/m),
  \qquad
  \E[N_r]=b,
  \qquad
  \operatorname{Var}(N_r)\leq b.
\]
Let $\theta_r\in[b]$ be the positive child in copy $r$, and put $q:=\lfloor b/2\rfloor$.
Chebyshev's inequality gives, for $b\geq8$,
\[
  \Pp(N_r\geq q)
  \geq1-\Pp\bigl(|N_r-b|\geq b/2\bigr)
  \geq1-\frac4b
  \geq\frac12.
\]

On the event $\{N_r\geq q\}$, the first $q$ within-copy child indices are independent uniform
draws from the same $b$ children.  Let $U_r^{(q)}$ be the number of their first occurrences that
come no later than the first occurrence of the positive child.  For a child $c$, let $\tau_c$
be its first occurrence among these draws, with $\tau_c=\infty$ if it does not occur.  By
symmetry, for every $c\neq\theta_r$,
\[
  \Pp(\tau_c\leq q,\ \tau_c<\tau_{\theta_r}\mid N_r\geq q)
  =\frac12\left(1-\left(1-\frac2b\right)^q\right).
\]
Indeed, this probability is
\[
  \sum_{t=1}^q\left(1-\frac2b\right)^{t-1}\frac1b.
\]
The positive child itself contributes when $\tau_{\theta_r}\leq q$.  Because the copy count is
independent of the within-copy child indices,
\[
  \E[U_r^{(q)}\mid N_r\geq q]
  =
  1-\left(1-\frac1b\right)^q
  +
  \frac{b-1}{2}\left(1-\left(1-\frac2b\right)^q\right)
  \geq \frac b4 .
\]
For $q=\lfloor b/2\rfloor$ and $b\geq8$, we have $2q/b\geq8/9$, and hence
\[
  1-\left(1-\frac2b\right)^q
  \geq 1-e^{-2q/b}
  \geq 1-e^{-8/9}
  >\frac47.
\]
Thus the second term alone is greater than $2(b-1)/7\geq b/4$.  Before any of these first
occurrences, at most $q-1<b-1$ children have been ruled out, so at least two possible positive
children remain.  Thus all $U_r^{(q)}$ counted rounds are uncertain.

Since $\Pp(N_r\geq q)\geq1/2$, the expected number of uncertain rounds contributed by active
copy $r$ is at least
\[
  \Pp(N_r\geq q)\,
  \E[U_r^{(q)}\mid N_r\geq q]
  \geq \frac b8.
\]
Summing over the $m$ active copies shows that every length-$mb$ block contains at least
$mb/8$ uncertain rounds in expectation, and hence the $L$ blocks contain at least $T_0/8$
in expectation.  This step uses only linearity; the copywise counts need not be independent.
Conditional on the learner's information after it observes the current child on an uncertain
round, symmetry leaves the positive child uniform among the $s$ children not yet ruled out,
where $2\leq s\leq b$.  Predicting zero errs with probability $1/s\geq1/b$, while predicting
one errs with probability $1-1/s\geq1/2\geq1/b$.  Thus every non-abstaining prediction on an
uncertain round has conditional mistake probability at least $1/b$.

Abstention can avoid the mistake on an uncertain round, so subtract the abstentions on those rounds
and upper bound their number by $N_\bot$.  We obtain
\begin{equation}
  \E[\MisErr]
  \geq\frac1b\left(\frac{T_0}{8}-\E[N_\bot]\right).
  \label{eq:lower-uncertain-tradeoff}
\end{equation}

\paragraph{Step 4: the budget tradeoff.}
Now set
\[
  b:=\left\lceil\frac{16A}{m}\right\rceil,
  \qquad
  L:=\left\lfloor\frac{T}{mb}\right\rfloor.
\]
Because $m\leq A$ and $A\leq T/32$,
\[
  b\geq16,
  \qquad
  b\leq\frac{32A}{m},
  \qquad
  mb\leq T,
  \qquad
  L\geq\frac{T}{2mb}.
\]
In particular, $T_0=Lmb\in[T/2,T]$.  As in Step~2, pad the remaining $T-T_0$ rounds with
corrupted repetitions of the first active tree's root.

If some fixed member of the hard family has $\E[\AbsErr]>A$, the first alternative of the theorem
holds.  Otherwise every member has expected clean abstentions at most $A$, so the uniform mixture
also satisfies $\E[\AbsErr]\leq A$.  Equation \eqref{eq:lower-abstention-identity} then gives
$\E[N_\bot]\leq AL$.  Substitute this into \eqref{eq:lower-uncertain-tradeoff} and use
$T_0=Lmb$:
\begin{align*}
  \E[\MisErr]
  &\geq L\left(\frac m8-\frac Ab\right)
   \geq\frac{Lm}{16}\\
  &\geq\frac{T}{32b}
   \geq\frac{mT}{1024A}\\
  &=\frac1{1024}\min\left\{\frac{dT}{A},\,T\right\}.
\end{align*}
Therefore some fixed member of the family forces the stated expected mistake lower bound.

For the total-cost consequence, suppose $1\leq d\leq T/1024$ and choose the integer budget
$A:=\lfloor\sqrt{dT}\rfloor$.  Then $1\leq A\leq T/32$, $A\geq\sqrt{dT}/2$, and
$dT/A\leq T$.  The tradeoff gives either
$\E[\AbsErr]>A\geq\sqrt{dT}/2$ or
\[
  \E[\MisErr]
  \geq\frac1{1024}\frac{dT}{A}
  \geq\frac1{1024}\sqrt{dT},
\]
which proves the displayed total-cost conclusion.
\end{proof}

\section{Unlabeled History Potential under Censored Feedback}
\label{app:censored-age-restarts}

In this section, we prove Theorem~\ref{thm:censored-age-restarts}. We prove the misclassification and abstention bounds separately in the following two sections.

\subsection{Mistake accounting and the number of active constraints}
\label{app:ar-mistakes}

\begin{lemma}[Misclassification bound]
 \label{lem:ar-mistakes}
 On every realizable interaction,
 Algorithm~\ref{alg:censored-age-restarts} satisfies
 \eqref{eq:ar-mistakes}. Moreover, it makes at most
 \begin{equation*}
  K_T:=\left\lceil
       \frac{4d\log(T+1)+2\beta}{\log(4/3)}\right\rceil
 \end{equation*}
 mistakes in each block. Every current version space is consequently
 specified by at most $K_T$ truthful constraints from the current block.
\end{lemma}

\begin{proof}
We first note that since every shattered set has size at most $d$ and weights are bounded by $\lambda_{u,t}\leq T+1$ for any $t\leq T$, we have for any version spaces $V,V'\subseteq\Fcal$
 \begin{equation*}
  1\le\widehat\Theta^u_t(V;U_t)
  \le \sum_{k=0}^d \binom{T}{k} (T+1)^k \leq (T+1)^{2d} \leq (T+1)^{2d}\widehat\Theta^u_{t-1}(V';U_{t-1}),
 \end{equation*}
 provided that $V$ and $V'$ are non-empty (realizable).

We now denote by $V_t$ the version space at the end of round $t$, and by $m_t$ the indicator whether a mistake was made at that round. If there was no restart of the version space at time $t$, one has
\begin{align*}
    \widehat\Theta^u_t(V_t;U_t)\leq (1+\beta)\widehat\Theta^u_t(V_t;U_{t-1})&\leq (1+\beta)\paren{\frac{3}{4}}^{m_t}\widehat\Theta^u_t(V_{t-1};U_{t-1})\\
    &\le (1+\beta)\left(1+\frac\beta d\right)^d \paren{\frac{3}{4}}^{m_t}\widehat\Theta^u_{t-1}(V_{t-1};U_{t-1}) \\
    &\leq e^{2\beta}\paren{\frac{3}{4}}^{m_t}\widehat\Theta^u_{t-1}(V_{t-1};U_{t-1}).
\end{align*}In the first and third inequalities we used the same arguments as in the non-censored case: appending a new point $x_t$ with weight $\beta$ increases the potential by at most a factor $1+\beta$ and $\lambda_{u,t}\leq(1+\beta/d)\lambda_{u,t-1}$ for all $u\leq t-1$.

In particular, denoting by $r$ the number of restarts and $M$ the total number of mistakes. Since the potential is always between $[1,(T+1)^{2d}]$, every restart inflates the potential by at most $(T+1)^{2d}$, hence
\begin{equation*}
    1\leq \widehat\Theta^u_T(V_T;U_T)\leq (T+1)^{2dr}e^{2\beta T}\paren{\frac{3}{4}}^M \leq e^{4\beta T}\paren{\frac{3}{4}}^M.
\end{equation*}
In the last inequality we used the fact that $r\leq T/W_\beta \leq \beta T/(d\log(T+1))$. Rearranging gives the desired misclassification bound. 

Next, within each block that has length at most $W_\beta$, there is at most $1$ restart (at the beginning). Hence the same arguments as above show that the corresponding number of mistakes $m$ satisfies
\begin{equation*}
    1\leq (T+1)^{2d} e^{2\beta W_\beta}\paren{\frac{3}{4}}^m \leq e^{2}(T+1)^{4d}\paren{\frac{3}{4}}^m.
\end{equation*}
Rearranging proves the desired bound on the number $m$ of mistakes per block.
\end{proof}

\subsection{Uniform concentration with capped weights}
\label{app:ar-uniform}

We fix the optimal function $f^\star$. For any subset $I\subseteq[T]$ we use the notation $w_t(I):=\prod_{u\in I}\lambda_{u,t}$ for the weight of $I$. For any set $C\subseteq[T]$ we also define the version space corresponding to mistakes on rounds in $C$ via
\begin{equation*}
    V_C:=\set{f\in\Fcal:\forall s\in C: f(x_s)=f^\star(x_s)}.
\end{equation*}
In particular, each version space $V_C$ is non-empty since it contains $f^\star$. For any non-empty version space $V\subseteq\Fcal$ we then define
\begin{equation*}
     a_t(U,V,x):=\mathbf 1\left\{
 \min_{b\in\{0,1\}}\widehat\Theta^u_t(V^{x\to b};U)
 >\frac34\widehat\Theta^u_t(V;U)\right\},
\end{equation*}
which is the indicator of whether round $t$ led to an abstention of the algorithm, provided the unlabeled history is $U$, the current version space is $V$,  and the new test point is $x$.

We start by relating the probability of an abstention to comparisons between weighted sums and potentials, similar to the replay-branch decomposition \cref{lem:empirical-pointwise}. 

\begin{lemma}\label{lemma:abstention_implies_large_extension_censored}
    For any $t\in[T]$, non-empty version space $V\subseteq\Fcal$ and unlabeled history $U$,
    \begin{equation*}
        \set{a_t(U,V,x)=1} \subseteq \set{ 2\sum_{I\subseteq[t-1]} w_t(I)\1[V\triangleright (x_I,x)] \geq \widehat\Theta^u_t(V;U)} .
    \end{equation*}
\end{lemma}
\begin{proof}
    As in the non-censored case, for every $I$ and $x$, the one-point shattering inequality is
 \[
  \mathbf 1\{V^{x\to0}\triangleright x_I\}
   +\mathbf 1\{V^{x\to1}\triangleright x_I\}
  \le\mathbf 1\{V\triangleright x_I\}
   +\mathbf 1\{V\triangleright(x_I,x)\}.
 \]
 Multiplying by the weight $w_t(I)$ and summing over the subsets $I$ shows that if an abstention is triggered at round $t$ ($a_t(U,V,x)=1$):
 \begin{equation*}
     \frac{1}{2}\widehat\Theta^u_t(V;U) \leq \widehat\Theta^u_t(V^{x\to 0};U)+\widehat\Theta^u_t(V^{x\to 1};U) - \widehat\Theta^u_t(V;U) \leq \sum_{I\subseteq[t-1]}
         w_t(I)\mathbf 1\{V\triangleright(x_I,x)\}.
 \end{equation*}
 This ends the proof.
\end{proof}

We are now ready to bound the probability of a bad abstention on a clean round $t$. We denote by $J:=\{j<t:c_j=0\}$ the set of previous clean rounds and by $i=|J|$ their number. We also fix an oblivious corruption schedule and all injected
contexts. For convenience, we introduce the quantity
\begin{equation*}
    h_j:=\beta\paren{1+\frac{\beta j}{d}}, j\geq 1,
\end{equation*}
which lower-bounds the weight of the $j$-th previous clean sample before $t$.

\begin{lemma}\label{lemma:proba_abstention_censored_bound}
    Fix $\delta\in(0,1)$ and a positive integer $k\leq i-d$. We introduce
    \begin{equation*}
        H:=\sum_{j=1}^{i-d-k}h_j \quad\text{and}\quad h:=6h_i ((k+d)\log(T+1)+\log(1/\delta)).
    \end{equation*}
    Then, with probability at least $1-\delta$, simultaneously over all subsets $C\subseteq[t-1]$ with $|C|\leq k$, one has
 \begin{equation*}
     \mu\set{x\in\Xcal:a_t(U_{t-1},V_C,x)=1} \leq \frac{4(d+h)}{H},
 \end{equation*}
 where the probability is taken only over the clean samples on clean rounds before $t$.
In particular, with $\delta=1/T^2$, if $i\geq 4(d+k)$ we have for some universal constant $c_0>0$, for any such subset $C\subseteq[t-1]$,
\begin{equation*}
    \mu\set{x\in\Xcal:a_t(U_{t-1},V_C,x)=1} \leq c_0\paren{\frac{d}{\beta i}}^2+c_0\frac{(k+d)\log(T+1)}{i}
\end{equation*}
\end{lemma}
\begin{proof}
    For any subset $I\subseteq[t-1]$ we introduce
    \begin{equation*}
        Z_{t,C,I}:=\sum_{j\in J\setminus(C\cup I)}\lambda_{j,t}\1[V_C \triangleright (x_I,x_j)]\quad \text{and} \quad q_{t,C,I}:=\Pbb_{x\sim\mu}[V_C\triangleright(x_I,x)].
    \end{equation*}

    We first lower-bound $Z_{t,C,I}$ as a function of $q_{t,C,I}$. For convenience, we introduce
    \begin{equation*}
        \bar\lambda_{j,t}:=\min(\lambda_{j,t},h_i),\; j\in J\setminus(C\cup I).
    \end{equation*}
    Conditionally on $x_{C\cup I}$, the variables $\1[V_C\triangleright(x_I,x_j)]$ for $j\in J\setminus(C\cup I)$ are i.i.d.\ Bernoulli variables of parameter $q_{t,C,I}$. Hence, Bernstein's inequality implies that with probability at least $1-\delta$ conditionally on $x_{C\cup I}$,
    \begin{align*}
        Z_{t,C,I}\geq \sum_{j\in J\setminus(C\cup I)} \bar\lambda_{j,t}\1[V_C\triangleright(x_I,x_j)] 
        &\geq \frac{1}{2} q_{t,C,I} \sum_{j\in J\setminus(C\cup I)} \bar\lambda_{j,t}-6h_i\log(1/\delta)\\
        &\geq \frac{1}{2}q_{t,C,I} \sum_{j=1}^{|J\setminus(C\cup I)|}h_j -6h_i\log(1/\delta). 
    \end{align*}
    In the first inequality we used the definition of $\lambda_{j,t}$.
    Hence, taking the union bound over all subsets $I,C\subseteq[t-1]$ with $|I|\leq d$ and $|C|\leq k$ we obtain simultaneously over all subsets $C\subseteq[t-1]$ of size at most $k$, with probability at least $1-\delta$:
    \begin{equation}\label{eq:bound_weighted_sum_Z}
         \sum_{\substack{I\subseteq[t-1]\\V_C\triangleright x_I}}
       w_t(I)Z_{t,C,I} \geq \frac{H}{2}\sum_{\substack{I\subseteq[t-1]\\V_C\triangleright x_I}}
       w_t(I)q_{t,C,I} -  h \widehat\Theta^u_t(V_C;U_{t-1}).
    \end{equation}
    Let $\mathcal E_t$ be the event on $U_{t-1}$ on which
\cref{eq:bound_weighted_sum_Z} holds simultaneously for all
$C\subseteq[t-1]$ with $|C|\leq k$.
Then $\Pbb(U_{t-1}\in\mathcal E_t)\geq1-\delta$.
    On the other hand, we note that
     \begin{align*}
  \sum_{\substack{I\subseteq[t-1]\\V\triangleright x_I}}
       w_t(I)Z_{t,C,I}
       &\le
    \sum_{\substack{I\subseteq[t-1]\\V\triangleright x_I}}
      w_t(I)\sum_{j\in[t-1]\setminus I}
      \lambda_{j,t}\,
      \mathbf 1[V_C\triangleright(x_I,x_j)]\\
  &=\sum_{\substack{J\subseteq[t-1]\\V\triangleright x_J}}
       |J|w_t(J)
    \le d\widehat\Theta^u_t(V_C;U_{t-1}),
 \end{align*}
 where in the last inequality we used the fact that each shattered set in $V_C$ has at most $d$ elements.
 
 Fix any $U_{t-1}\in\mathcal E_t$ and any
$C\subseteq[t-1]$ with $|C|\leq k$. Together with
\cref{eq:bound_weighted_sum_Z},
\cref{lemma:abstention_implies_large_extension_censored} implies that
\begin{align*}
\Ebb_{x\sim\mu}[a_t(U_{t-1},V_C,x)]
&\leq \Pbb_{x\sim\mu}\left[
\frac{4(d+h)}{H}
\sum_{\substack{I\subseteq[t-1]\\V_C\triangleright x_I}}
w_t(I)\1[V_C\triangleright(x_I,x)]
\geq
\sum_{\substack{I\subseteq[t-1]\\V_C\triangleright x_I}}
w_t(I)q_{t,C,I}
\right]\\
&\leq \frac{4(d+h)}{H}
\Ebb_{x\sim\mu}\left[
\frac{
\sum_{\substack{I\subseteq[t-1]\\V_C\triangleright x_I}}
w_t(I)\1[V_C\triangleright(x_I,x)]
}{
\sum_{\substack{I\subseteq[t-1]\\V_C\triangleright x_I}}
w_t(I)q_{t,C,I}
}
\right]\\
&=\frac{4(d+h)}{H}.
\end{align*}
The equality follows from the definition of $q_{t,C,I}$.
With at least $1-\delta$ probability over $U_{t-1}$, this holds simultaneously over all eligible $C$.
Simplifying the expression proves the second claim.

\end{proof}\cref{lemma:proba_abstention_censored_bound} bounds the probability of a bad abstention uniformly over the potential version spaces. During the run of the algorithm, these version spaces are defined by at most $K_T$ constraints as per \cref{lem:ar-mistakes}. Putting these together then gives the desired abstention bound.

\begin{proof}[Proof of Theorem~\ref{thm:censored-age-restarts}]
 Lemma~\ref{lem:ar-mistakes} proves the misclassification bound.
 For abstentions, we condition on $f^\star$, injected contexts, and clean rounds, which we denote $s_1<\ldots<s_N$. We apply Lemma~\ref{lemma:proba_abstention_censored_bound} with $k=K_T$ to each $i$-th clean round for $i\geq 4(d/\beta+K_T)$, which gives
 \begin{align*}
     \Ebb[\operatorname{AbsErr}] &= \Ebb\sqb{\sum_{i=1}^N a_{s_i}(U_{s_i-1},V_{s_i},x_{s_i})} \\
     &\leq 4\paren{\frac{d}{\beta}+K_T}+\sum_{i\geq 4(d/\beta+K_T)}\paren{c_0\paren{\frac{d}{\beta i}}^2 + c_0\frac{(K_T+d)\log(T+1)}{i}+\frac{1}{T^2}}\\
     &\leq c_1\paren{\frac{d}{\beta} + d\log^3(T+1))},
 \end{align*}
 for some universal $c_1>0$. This ends the proof.
\end{proof}

\section{Proofs for Learning Against Adaptive Adversaries}
\label{sec:ac-proofs}

The proof follows the three parts of the construction. First, in \cref{subsec:preliminaries_compression}, we
show that reduction dimension gives short witness descriptions and
that the compressed rule appropriately approximates the original scores $a_b(x)$. In particular, this will give the desired misclassification error bound for the reference process. We turn to bounding its abstention error in \cref{subsec:proof_abstention_reference_exp}. We then prove the guarantee for the aggregation of experts in \cref{app:adaptive-direct-aggregation}, and complete the proofs of the performance of our algorithms in \cref{subsec:final_proofs}.

\subsection{Preliminaries on the witness compression}
\label{subsec:preliminaries_compression}

We first check that the reduction dimension can be used to bound the VC dimension of witness classes.

\begin{lemma}[VC dimension of the witness class]
\label{lem:ac-witness-vc}
Fix $\ell\geq 0$. On the disjoint union $\Xcal^{\leq d}:=\bigcup_{k=0}^d\mathcal X^k$, the function class
\begin{equation*}
    \Gcal:=\{Z\in\Xcal^{\leq d}\mapsto \1[W\triangleright Z]: W\in\mathcal R_\ell\}
\end{equation*}
has VC dimension at most $C_0D\log(2dD)$ for a universal
constant $C_0$, where $D$ is the $\ell$-reduction dimension of $\Fcal$.
\end{lemma}

\begin{proof}
It suffices to consider $r\geq D$. Fix $r$ tuples in $\Xcal^{\leq d}$. All $r$ shattering
indicators are determined by the trace of $W$ on the union of their
coordinates, which has $n\leq rd$ points. As $rd\geq D$, the reduction-dimension
assumption bounds the number of possible indicator vectors by
$(rd)^D$. If the $r$ tuples are shattered by $\mathcal G$, then
$2^r\leq (rd)^D$. In particular, we must have $r\leq C_0D\log(2dD)$ for some universal $C_0$.
\end{proof}

We are now ready to check that short witnesses are well-defined.

\begin{lemma}[Existence of a short witness list]
\label{lem:ac-compression} There is a universal constant $C_s>0$ such that the following holds. Let $D_T$ be the $\ell_T$-reduction dimension of $\Fcal$.
Fix any bank $S_1,\ldots,S_m$ and any nonempty $V\in\mathcal R_{\ell_T-1}$. Let $s=\ceil{C_sD_T\log(2dD_T)}$. There is a list of $s$ witness tuples $Z_1,\ldots,Z_s$, each using at most $d$ bank
coordinates among $S_1,\ldots,S_m$, such that
\[
 \sup_{x\in\mathcal X, b\in\{0,1\}}
 \left|\frac1s\sum_{u=1}^s \1[V^{x\to b}\triangleright Z_u]-a_b(x;V)\right|
 \le\frac1{64}.
\]
\end{lemma}

\begin{proof}
Fix $(x,b)$. Since any shattered set in $V^{x\to b}$ must also be shattered in $V$, one has
\begin{align*}
    \Pbb_{Z\sim\nu_V}[V^{x\to b}\triangleright Z] =\frac1m\sum_{j=1}^m
   \sum_{I\subseteq[N]}
   \frac{T^{|I|}\1[V\triangleright S_{j,I}]}
   {\binom N{|I|}\Theta^{S_j}(V)} \1[V^{x\to b}\triangleright S_{j,I}]=a_b(x;V).
\end{align*}
Apply uniform
convergence to the VC class in Lemma~\ref{lem:ac-witness-vc}, at
accuracy $1/64$ and success probability $1/2$, under the fixed
distribution $\nu_V$ (e.g. see \cite[Theorem 6.8]{shalev2014understanding}). This precisely ensures the existence of the desired witnesses $Z_1,\ldots,Z_s$ such that 
\begin{equation*}
    \sup_{W\in\Rcal_{\ell_T}}
 \left|\frac1s\sum_{u=1}^s \1[W\triangleright Z_u]-\mathbb E_{Z\sim\nu_V} [W\triangleright Z]\right|
 \le\frac1{64}.
\end{equation*}
It then suffices to check that for any $(x,b)\in\Xcal\times\{0,1\}$ we have $V^{x\to b}\in\mathcal R_{\ell_T}$.
\end{proof}

For a nonempty class $W\subseteq\Fcal$ and a bank $S\subseteq\Xcal$, we define
\begin{equation*}
 g_W(S,x)=\mathbf 1\left\{
   \Theta^S(W^{x\to0})+\Theta^S(W^{x\to1})
                       >\frac54\Theta^S(W)\right\}.
 \label{eq:ac-one-block-indicator}
\end{equation*}
Next, we check that using the threshold $3/4$ on the approximate score $\widehat a_b(x;V)$ to trigger an abstention has adequate properties in terms of the estimated weighted potentials.

\begin{lemma}
\label{lem:ac-margins}
Fix $V$ and a new test point $x$. For the reference process:
\begin{enumerate}
 \item On a mistaken prediction with true label $y$, $(\prod_{j\in[m]}\Theta^{S_j}(V^{x\to y}))^{1/m} < \frac{15}{16} (\prod_{j\in[m]}\Theta^{S_j}(V))^{1/m}$.
 \item On an abstention, $|\set{j\in[m]: g_V(S_j,x)=1}|>m/6$.
\end{enumerate}
In particular, the reference process makes at most $M_T$ mistakes.
\end{lemma}

\begin{proof}
On a mistaken prediction, we have by construction $\widehat a_y(x;V)\leq 3/4$. Hence, using the arithmetico-geometric inequality:
\[
 \left(\prod_{j=1}^m \frac{\Theta^{S_j}(V^{x\to y})}{\Theta^{S_j}(V)}\right)^{1/m}
 \le\frac1m\sum_{j=1}^m \frac{\Theta^{S_j}(V^{x\to y})}{\Theta^{S_j}(V)}
 =a_y(x;V) \leq \widehat a_y(x;V) + \frac{1}{64}\le\frac34+\frac1{64}<\frac{15}{16}.
\]

Next, for any nonempty class $W$ and block $S$,
\[
 1\le\Theta^S(W)\le\sum_{k=0}^d T^k\le(1+T)^d.
\]
Let $P(V)=(\prod_j\Theta^{S_j}(V))^{1/m}$. After $r$
mistakes with valid witness lists, $1\le P(V)\le(1+T)^d(15/16)^r$.
This precisely implies the desired bound $r\leq M_T$ on the number of mistakes.

As a result, the version space $V_t$ of the reference process at the beginning of any iteration $t\geq t_0$ has at most $M_T=\ell_T-1$ constraints: $V\in\Rcal_{\ell_T-1}$. Then, \cref{lem:ac-compression} implies that $|a_b(x;V)-\widehat a_b(x;V)|\leq 1/64$ for both labels $b\in\{0,1\}$. In turn, on an abstention we must have for both labels $a_b(x;V)>3/4-1/64>11/16$. Therefore,
\begin{align*}
    \frac{11}{8}<a_0(x;V)+a_1(x;V) &= \frac{1}{m}\sum_{j\in[m]} \frac{\Theta^{S_j}(V^{x\to0})}{\Theta^{S_j}(V)}+\frac{\Theta^{S_j}(V^{x\to1})}{\Theta^{S_j}(V)}\\
    &\leq \frac{1}{m}\sum_{j\in[m]}\paren{2g_V(S_j,x) + \frac{5}{4}(1-g_V(S_j,x))}.
\end{align*}
Rearranging gives the desired bound on the number of blocks $j\in[m]$ for which $g_V(S_j,x)=1$. This ends the proof.
\end{proof}

\subsection{A Leave-One-Out Bound for Clean Abstentions}
\label{subsec:proof_abstention_reference_exp}
We start by proving a bound on the expectation of the indicator $g_W(S,x)$ when the elements of the bank $S$ and the test-point $x$ are exchangeable, using a leave-one-out argument.

\begin{lemma}[One-block leave-one-out bound]
\label{lem:ac-loo}
Fix any nonempty class $W\subseteq\mathcal F$ and a tuple $R=(z_1,\ldots,z_{N+1})\in \Xcal^{N+1}$. Let $i$ be a random uniform index within $[N+1]$. Then,
\[
 \mathbb E_i[g_W(R_{-i},z_i)]\le\frac4T+\frac d{N+1}.
\]
\end{lemma}

\begin{proof}
Define for each $i\in[N+1]$ the quantity
\[
E_i=\sum_{I\subseteq[N+1]\setminus\{i\}}
\frac{T^{|I|}}{\binom{N}{|I|}}
\1[W\triangleright(z_i,R_I)].
\]
As in the previous sections, if either branch $W^{z_i\to b}$
shatters a tuple, then $W$ shatters it; if both branches shatter
it, then $W$ shatters its extension by $z_i$:
\[
\Theta^{R_{-i}}(W^{z_i\to0})
+\Theta^{R_{-i}}(W^{z_i\to1})
\leq \Theta^{R_{-i}}(W)+E_i.
\]
In particular, if $g_W(R_{-i},z_i)=1$, the previous inequality
implies $E_i>\Theta^{R_{-i}}(W)/4$.

Next, we compare $\Theta^R(W)$ to $\Theta^{R_{-i}}(W)$,
the potential where one sample was removed. Splitting terms
according to whether they contain $i$ gives
\begin{align*}
\Theta^R(W)
&\leq \Theta^{R_{-i}}(W)
+\sum_{\substack{I\subseteq[N+1]\\i\in I}}
\frac{T^{|I|}}{\binom{N+1}{|I|}}
\1[W\triangleright R_I]\\
&\overset{(i)}{\leq}\Theta^{R_{-i}}(W)
+\frac{d}{N+1}
\sum_{\substack{I\subseteq[N+1]\\i\in I}}
\frac{T^{|I|}}{\binom{N}{|I|-1}}
\1[W\triangleright R_I]=\Theta^{R_{-i}}(W)+\frac{dT}{N+1}E_i.
\end{align*}
In $(i)$ we used the fact that shattered sets in $W\subseteq\mathcal F$
have size at most $d$.
Therefore, if $g_W(R_{-i},z_i)=1$, we have
\[
E_i>\frac{\Theta^R(W)}{4+Td/(N+1)}.
\]

On the other hand, we can bound the sum of the quantities $E_i$
by counting each nonempty shattered subset once for each
distinguished coordinate:
\begin{align*}
\sum_{i=1}^{N+1}E_i
&=\sum_{\varnothing\neq I\subseteq[N+1]}
|I|\frac{T^{|I|-1}}{\binom{N}{|I|-1}}
\1[W\triangleright R_I]\\
&=\frac{N+1}{T}
\sum_{\varnothing\neq I\subseteq[N+1]}
\frac{T^{|I|}}{\binom{N+1}{|I|}}
\1[W\triangleright R_I]\\
&=\frac{N+1}{T}\bigl(\Theta^R(W)-1\bigr)
\leq \frac{N+1}{T}\Theta^R(W).
\end{align*}
Together with the previous lower bound on $E_i$ when
$g_W(R_{-i},z_i)=1$, this shows that
\[
\left|\{i\in[N+1]:g_W(R_{-i},z_i)=1\}\right|
\leq \frac{N+1}{T}\left(4+\frac{Td}{N+1}\right)
=\frac{4(N+1)}{T}+d.
\]
Dividing by $N+1$ ends the proof.
\end{proof}

Essentially, \cref{lem:ac-loo} shows that each block of size $N$ leads to a clean abstention error rate of order $O(1/N)$ for a fixed class $W$. Using the reduction dimension, we lift this abstention error rate for all classes $W\in\Rcal_{\ell_T}$ constructed using at most $\ell_T$ constraints. This is sufficient since all version spaces arising during the run of the reference expert have at most $\ell_T$ constraints. Using a few blocks $m$ precisely ensures a sufficiently small failure probability so that we can apply uniform concentration over all such classes in $\Rcal_{\ell_T}$.

\begin{lemma}[Uniform block certificate]
\label{lem:ac-uniform-certificate}
There is a universal constant $C_m>0$ such that the following holds. Let $\delta\in(0,1)$, $N\geq d$, $\ell\geq 0$, and suppose that $m\ge C_m(D\log(2ND)+\log\frac2\delta)$ where $D$ is the $\ell$-reduction dimension of $\Fcal$. Sample independently $m$ blocks $S_1,\ldots,S_m\sim\mu^N$ of $N$ samples each. Next, define
\begin{equation*}
    Q_\ell(S_1,\ldots,S_m):=\sup_{\emptyset\neq W\in\Rcal_\ell} \mu\set{x:\frac1m\sum_{j=1}^m g_W(S_j,x)>\frac16}.
\end{equation*}
Then,
\begin{equation*}
 \Ebb_{S_1,\ldots,S_m}[Q_\ell(S_1,\ldots,S_m)]\le\frac{32}{T}+\frac{8d}{N+1}+\delta.
 \label{eq:ac-Q-bound}
\end{equation*}
\end{lemma}

\begin{proof}
On $r$ pairs $(S,x)$, the indicator vector
$(g_W(S,x))$ for classes $W\in\Rcal_\ell$ is determined by the trace of such $W$ on at most
$r(N+1)$ coordinates. The counting argument from
Lemma~\ref{lem:ac-witness-vc} therefore bounds the VC dimension
of this class by $C_1D\log(2ND)$ for some universal $C_1>0$. Then, uniform convergence at the constant gap $1/24$ shows that for an appropriate choice of the constant $C_m$, for each fixed $x_0\in\Xcal$, with probability at least $1-\delta$ over $S_1,\ldots,S_m$,
\begin{equation*}\label{eq:unif_concen_guarantee_subsets}
    \sup_{W\in\Rcal_\ell}\abs{\frac{1}{m}\sum_{j\in[m]}g_W(S_j,x_0)-\Ebb_{S\sim\mu^N}[g_W(S,x_0)]} < \frac{1}{24}.
\end{equation*}
Additionally taking expectation over $x\sim\mu$ gives,
\[
 \mathbb E_{S_1,\ldots,S_m}\sqb{\mu(\Bcal)
 }\le\delta \quad\text{where} \quad \Bcal:=\left\{x:\exists W\in\Rcal_\ell, \Ebb_{S\sim\mu^N}[g_W(S,x)]\le\frac18\text{ and }
             \frac1m\sum_{j\in[m]} g_W(S_j,x)>\frac16\right\}
\]

Next, since i.i.d.\ samples from $\mu$ are exchangeable, Lemma~\ref{lem:ac-loo} shows that for any non-empty class $W$,
\begin{equation*}
    \Ebb_{S\sim\mu^N,x\sim\mu}[g_W(S,x)]\leq \frac{4}{T}+\frac{d}{N+1}.
\end{equation*}
In particular, Markov's inequality gives for every non-empty class $W\in\Rcal_\ell$,
\begin{align*}
    \mu\set{\frac{1}{m}\sum_{j\in[m]}g_W(S_j,x)>\frac{1}{6}} \leq \mu\paren{\Bcal\cup\set{x:\Ebb_{S\sim\mu^N}[g_W(S,x)]>\frac18}} \leq \frac{32}{T}+\frac{8d}{N+1} +\mu(\Bcal).
\end{align*}
Taking the supremum over all non-empty $W\in\Rcal_\ell$ then the expectation over $S_1,\ldots,S_m$ ends the proof, since $\Ebb[\mu(\Bcal)]\leq\delta$.
\end{proof}

We are now ready to bound the abstention error of the reference expert.

\begin{proof}[Proof of Proposition~\ref{prop:performance_reference_expert_no_restart}]
We already bounded the misclassification error within \cref{lem:ac-margins}. We now focus on the abstention error. To simplify the analysis, we view all clean samples as coming from an infinite sequence of i.i.d.\ draws $Z_1,Z_2,\ldots \overset{iid}{\sim}\mu$: the sample observed on the $i$-th clean round is $Z_i$. 

Using the first $mN$ elements of the sequence $(Z_i)_{i\geq 1}$ we construct the $m$ blocks $S_1,\ldots,S_m$. Importantly, we define these even if fewer than $mN$ clean rounds occur during the interaction. Now focus on any clean round $t\geq t_0$ (recall that $t_0$ is the time immediately after the $mN$-th clean round that occurred). For convenience, we denote by $V_t$ the version space at the beginning of round $t$ and by $\Hcal_t$ the history up to time $t$ included. Using \cref{lem:ac-margins}, we can bound the probability of a bad abstention on round $t$ by
\begin{align}
    \Pbb[\widehat y_t=\bot \mid \Hcal_{t-1},c_t=0] &\leq \Pbb\sqb{\frac{1}{m}\sum_{j\in[m]} g_{V_t}(S_j,x_t)>\frac16 \mid \Hcal_{t-1},c_t=0} \notag\\
    &=\mu\set{x:\frac{1}{m}\sum_{j\in[m]} g_{V_t}(S_j,x)>\frac16} \leq Q_{\ell_T}(S_1,\ldots,S_m). \label{eq:bound_abstention_one_round}
\end{align}
In the last inequality we used the fact that $V\in\Rcal_{\ell_T}$. Before time $t_0$, the reference expert always abstains, incurring abstention error at most $mN$. We next use the notation $c(T)=|\{t\geq t_0: c_t=0\}|$ for the number of clean samples after the start time $t_0$. Starting from $t_0$, the expected abstention error is then
\begin{align}
    \Ebb\sqb{\sum_{t\geq t_0}^T \1[\widehat y_t=\bot]\1[c_t=0]} &\leq \Ebb[c(T)Q_{\ell_T}(S_1,\ldots,S_m)]\label{eq:useful_bound_for_multi_phase}\\
    &\leq T\Ebb[Q_{\ell_T}(S_1,\ldots,S_m)]\notag \\
    &\leq \frac{8dT}{N+1}+32+\delta T,\notag
\end{align}
where in the last inequality we used the assumption on $m$ and \cref{lem:ac-uniform-certificate}.

Adding the two contributions to the expected abstention error proves the desired bound.
\end{proof}

\subsection{Direct Aggregation over Candidate Banks}
\label{app:adaptive-direct-aggregation}

Write $L:=|\mathcal C|$, $R:=1+\log L+(\log2)M$, and
$\gamma:=R/(2\kappa T)$.  We assume $R\le2\kappa T$, so that $\gamma\le1$.

\begin{proof}[Proof of \cref{prop:adaptive-direct-aggregation-guarantee}]
The comparison uses the total expert weight in two directions.  Learner mistakes give an upper
bound by contracting the total weight, while the weight of one expert gives a lower
bound and records rounds on which the learner abstains but that expert predicts.

Suppose first that the learner predicts and is wrong.  The true label has minimum aggregate mass
$P_{y_t}$, possibly tied, and this mass is at most $(1-\kappa)\Phi_t$.  The complementary
weight, consisting exactly of experts that predicted the learner's wrong label, is at least
$\kappa\Phi_t$.  Halving those weights gives
\[
  \Phi_{t+1}
  \le
  \left(1-\frac\kappa2\right)\Phi_t
  \le
  e^{-\kappa/2}\Phi_t.
\]

If the learner abstains, the weight predicting either label is less than $\kappa\Phi_t$.  The
total weight of experts that predict is therefore less than $2\kappa\Phi_t$, which gives
\[
  \Phi_{t+1}
  \le
  (1+2\gamma\kappa)\Phi_t
  \le
  e^{2\gamma\kappa}\Phi_t.
\]
On a correct learner prediction, the total weight cannot increase.

For an expert $c$, let
$D_c:=|\{t:\widehat y_t=\bot,\ \widehat y_{c,t}\ne\bot\}|$, and let
$\MisErr^{(c)}$ be the total number of its mistakes.
Since $\Phi_1=1$, iterating the one-round bounds and using at most $T$ learner abstentions gives
\[
  \Phi_{T+1}
  \le
  \exp\left(
    2\gamma\kappa T-\frac\kappa2\MisErr
  \right).
\]
On the other hand, the two multiplicative updates give the exact final weight of expert $c$:
\[
  w_{c,T+1}
  \geq
  \frac1L
  (1+\gamma)^{D_c}
  2^{-\MisErr^{(c)}}.
\]
Combining $w_{c,T+1}\le\Phi_{T+1}$ with these two bounds gives
\begin{equation}
  \frac\kappa2\MisErr
  +
  \log(1+\gamma)D_c
  \le
  \log L
  +
  (\log2)\MisErr^{(c)}
  +
  2\gamma\kappa T.
  \label{eq:adaptive-aggregate-comparison}
\end{equation}

If $\MisErr^{(c)}\le M$, the right side is at most $2R$.  Dropping the second term
on the left gives $\MisErr\le4R/\kappa$.  Dropping the first term instead and using
$\log(1+\gamma)\ge\gamma/2$ gives
$D_c\le4R/\gamma=8\kappa T$.
Finally, on every round in $U$ where the learner abstains, either expert $c$ also abstains or
the round contributes to $D_c$.  Thus the learner abstains on at most $D_c$ more rounds of
$U$ than expert $c$, which proves the simultaneous comparison.
\end{proof}

\subsection{Completing the Adaptive Theorem Proofs}
\label{subsec:final_proofs}

We start by proving the performance guarantee for the multi-phase reference expert \cref{prop:ac-reference}, then prove the main results \cref{thm:adaptive-full-feedback-direct,thm:adaptive-full-feedback-direct_no_phase}

\begin{proof}[Proof of Proposition~\ref{prop:ac-reference}]
On each phase $j$, the multi-phase reference expert coincides with the previous reference expert which starts at time $\tau_j$ and uses the parameters $m$ and $N_j$. Hence, \cref{prop:performance_reference_expert_no_restart} already shows that it makes at most $M_T$ mistakes on that phase. Since there are at most $J$ phases, this results in at most $JM_T$ misclassification error.

For the abstention error, the same arguments as in the proof of \cref{prop:performance_reference_expert_no_restart} (see \cref{eq:bound_abstention_one_round,eq:useful_bound_for_multi_phase}) show that
\begin{align*}
    \Ebb\sqb{\sum_{t=\tau_j}^{\tau_{j+1}-1} \1[\widehat y_t=\bot]\1[c_t=0]} &\leq \Ebb\sqb{\sum_{t=\tau_j}^{\tau_{j+1}-1}\1[c_t=0]Q_{\ell_T}(S_1^{(j)},\ldots,S_m^{(j)})}\\
    &\leq \min(n_{j+1}-n_j,T)\Ebb[Q_{\ell_T}(S_1^{(j)},\ldots,S_m^{(j)})] \\
    &\leq \frac{8d(n_{j+1}-n_j)}{N_j+1}+32+\delta T\leq 8dm + 32+\delta T,
\end{align*}
where $S_1^{(j)},\ldots,S_m^{(j)}$ denote the $N_j$-size clean-bank blocks used for phase $j$. Summing over at most $J$ phases and recalling that the multi-phase expert abstains for the first $md$ clean rounds, gives the desired expected abstention error bound.
\end{proof}

We are now ready to prove the main results---which is essentially putting together the reference expert guarantee \cref{prop:performance_reference_expert_no_restart,prop:ac-reference} and the direct aggregation guarantee \cref{prop:adaptive-direct-aggregation-guarantee}.

\begin{proof}[Proof of Theorem~\ref{thm:adaptive-full-feedback-direct}]
The cases $T=1$, $d>T$, or $D_T>T$ can be handled directly by either always abstaining ($0$ misclassification) or always predicting ($0$ abstention). We assume 
$T\ge2$ and $1\le d,D_T \le T$ from now on. We choose $\delta=1/T$ and $m=\left\lceil C_m\bigl[D_T\log(2TD_T)+\log(2T)\bigr]\right\rceil$. For the direct aggregation procedure, we use the parameter $\kappa=\epsilon/8$ and the family of experts $\Ccal$ obtained for any choice of start times $\tau_0<\ldots<\tau_{h-1}\leq T$ for $h\leq J=\ceil{\log_2 T}$ phases, and corresponding witness-index lists $(\mathcal Z_{j,r})_{j<h,0\leq r\leq M_T}$ which each contain $s=\ceil{C_s D_T\log(2dD_T)}$ tuples of at most $d$ indices in $[T]$. In particular, $L:=|\widetilde \Ccal| \leq (T+1)^{J+J(M_T+1)ds}$. We also use the parameter $\gamma=R/(2\kappa T)$, where $R:=1+\log L+(\log2)JM_T=O(d^2D_T\log^4 T)$. Note that if $R\geq 2\kappa T=\varepsilon T/4$, the desired misclassification bound can trivially be achieved by always predicting arbitrarily. We then suppose that $R\leq 2\kappa T$ from now.

By
Proposition~\ref{prop:ac-reference}, we can bound the misclassification and abstention errors $M_{\rm ref},A_{\rm ref}$ of the multi-phase reference expert by
\[
 M_{\rm ref}\le JM_T\leq Cd\log^2 T\quad\text{and}\quad
 \mathbb E A_{\rm ref}
 \le CdD_T\log^2 T.
\]
for some universal constant $C>0$. Then, by \cref{prop:adaptive-direct-aggregation-guarantee}, the direct aggregation guarantee gives
\[
 \operatorname{MisErr}\le 32R/ \varepsilon \quad\text{and} \quad
 \mathbb E\operatorname{AbsErr}
 \le \Ebb[A_{\rm ref}+8\kappa T] \leq \varepsilon T+CdD_T\log^2 T.
\]
This ends the proof.
\end{proof}

\begin{proof}[Proof of \cref{thm:adaptive-full-feedback-direct_no_phase}]
    As above, the cases $T=1$, $d>T$, or $D_T>T$ are immediate, so assume
$T\ge2$ and $1\le d,D_T\le T$. Choose
$\delta=1/T$, $m=\left\lceil C_m(D_T\log(2TD_T)+\log(2T))\right\rceil$, and
$N=\max\{d,\ceil{\sqrt{dT/m}}\}$.
If $md\geq T$, the trivial bound gives
\[
 \mathbb E A_{\rm ref}\leq T\leq\sqrt{dmT}.
\]
Otherwise, $md<T$, so $\sqrt{dT/m}>d$ and hence $N=\ceil{\sqrt{dT/m}}$. Consequently,
\[
 mN\leq\sqrt{dmT}+m\leq2\sqrt{dmT},
 \qquad
 \frac{dT}{N+1}\leq\sqrt{dmT}.
\]
Applying \cref{prop:performance_reference_expert_no_restart} with $\delta=1/T$, we obtain in either case
\[
 M_{\rm ref}\leq M_T,\qquad
 \mathbb E A_{\rm ref}\leq C_1\sqrt{dmT}
\]
for some universal $C_1>0$. For the direct aggregation procedure, we use $\kappa=\epsilon/8$, and the family of experts $\mathcal C$ obtained for any choice of start time
$t_0\in[T+1]$ and witness-index lists
$(\mathcal Z_r)_{0\le r\le M_T}$, each containing
$s=\lceil C_sD_T\log(2dD_T)\rceil$ tuples of at most $d$ indices
preceding $t_0$. In particular, $L:=|\Ccal|\le (T+1)^{1+(M_T+1)ds}$. We then pose $R:=1+\log L+(\log2)M_T
 =O\bigl(d^2D_T\log(2dD_T)\log^2(T+1)\bigr)$ and use the parameter $\gamma=R/(2\kappa T)$. As above, if $\gamma>1$, always predicting arbitrarily gives the desired bound. We now suppose $\gamma\leq 1$.
 
By \cref{prop:adaptive-direct-aggregation-guarantee} the direct
aggregation guarantee gives
\[
 \MisErr\le 32R/\epsilon\quad\text{and} \quad 
 \mathbb E\AbsErr
 \le \mathbb E [A_{\rm ref}+8\kappa T]
 \le \epsilon T+C\sqrt{dmT}.
\]
This ends the proof.
\end{proof}

\end{document}